\documentclass[11pt]{article}

\usepackage[final]{acl}

\usepackage{times}
\usepackage{latexsym}

\usepackage[T1]{fontenc}

\usepackage[utf8]{inputenc}

\usepackage{microtype}

\usepackage{inconsolata}

\usepackage{graphicx}

\usepackage{amsmath}
\usepackage{booktabs}
\usepackage{adjustbox}
\usepackage{multirow}

\usepackage{cleveref}
\usepackage{amsthm}

\usepackage{graphicx}
\usepackage{subcaption}
\usepackage{tcolorbox}
\tcbuselibrary{skins, breakable, listings}
\usepackage{listings}
\usepackage{amsfonts}
\usepackage{colortbl}
\usepackage{pifont}     %
\usepackage{makecell}

\usepackage{marvosym}
\usepackage{fontawesome5}
\usepackage{simpleicons}
\usepackage{twemojis}
\usepackage[dvipsnames]{xcolor}
\usepackage[inline]{enumitem}

\definecolor{mine_font}{RGB}{0, 128, 0}
\definecolor{tea_green}{RGB}{214, 234, 193}
\definecolor{hint_green}{RGB}{226,246,209}
\definecolor{Madang}{RGB}{190,235,159}
\definecolor{yellow_green}{RGB}{198,222,119}
\definecolor{link_water}{RGB}{221, 232, 250}
\definecolor{celestial_blue}{RGB}{52, 152, 219}
\definecolor{shakespeare}{RGB}{85, 154, 193}
\definecolor{buttermilk}{RGB}{255,242,174}
\definecolor{chardonnay}{RGB}{250,196,114}
\definecolor{rajah}{RGB}{253,180,98}
\definecolor{fog}{RGB}{213, 193, 234}
\definecolor{melon}{RGB}{254,191,181}
\definecolor{sundown}{RGB}{249, 180, 181}
\definecolor{mona_lisa}{RGB}{246,152,134}
\definecolor{salmon}{RGB}{242,131,107}
\definecolor{blue_x}{RGB}{142, 207, 201}
\definecolor{orange_x}{RGB}{255, 190, 122}

\definecolor{saltpan}{RGB}{238, 243, 232}
\definecolor{aqua_spring}{RGB}{232, 243, 232}
\definecolor{tea_green}{RGB}{214, 234, 193}
\definecolor{Madang}{RGB}{190,235,159}
\definecolor{fringy_flower}{RGB}{194, 234, 193}
\definecolor{aero_blue}{RGB}{193, 234, 213}
\definecolor{pixie_green}{RGB}{183,214,170}
\definecolor{french_pass}{RGB}{195,232,246}
\definecolor{ice_cold}{RGB}{169,232,220}
\definecolor{pale_turquoise}{RGB}{172,240,242}
\definecolor{cruise}{RGB}{179,226,205}
\definecolor{sail}{RGB}{163,205,235}
\definecolor{spindle}{RGB}{179,205,227}
\definecolor{link_water}{RGB}{221, 232, 250}
\definecolor{periwinkle}{RGB}{203,213,232}
\definecolor{zanah}{RGB}{220, 233, 213}
\definecolor{frostee}{RGB}{217, 231, 214}
\definecolor{opal}{RGB}{199, 221, 211}
\definecolor{jet_stream}{RGB}{188, 214, 210}
\definecolor{skeptic}{RGB}{153, 187, 167}
\definecolor{hint_green}{RGB}{226,246,209}
\definecolor{snow_flurry}{RGB}{230,245,201}
\definecolor{surf_crest}{RGB}{205,230,208}
\definecolor{yellow_green}{RGB}{198,222,119}
\definecolor{cream}{RGB}{255,255,204}
\definecolor{pale_prim}{RGB}{255,255,179}
\definecolor{spring_sun}{RGB}{242,243,195}
\definecolor{portafino}{RGB}{245,237,160}
\definecolor{buttermilk}{RGB}{255,242,174}
\definecolor{cream_brulee}{RGB}{255, 229, 151}
\definecolor{dairy_cream}{RGB}{254,226,189}
\definecolor{champagne}{RGB}{254,217,166}
\definecolor{chardonnay}{RGB}{250,196,114}
\definecolor{manhattan}{RGB}{226,180,125}
\definecolor{rajah}{RGB}{253,180,98}
\definecolor{early_dawn}{RGB}{252,243,218}
\definecolor{egg_shell}{RGB}{238, 234, 215}
\definecolor{selago}{RGB}{243, 232, 243}
\definecolor{quartz}{RGB}{219,223,238}
\definecolor{fog}{RGB}{213, 193, 234}
\definecolor{languid_lavender}{RGB}{222,203,228}
\definecolor{watusi}{RGB}{254,221,207}
\definecolor{coral_andy}{RGB}{243,204,205}
\definecolor{cosmos}{RGB}{248,209,210}
\definecolor{melon}{RGB}{254,191,181}
\definecolor{azalea}{RGB}{234, 193, 194}
\definecolor{beauty_bush}{RGB}{235, 185, 179}
\definecolor{sundown}{RGB}{249, 180, 181}
\definecolor{mona_lisa}{RGB}{246,152,134}
\definecolor{salmon}{RGB}{242,131,107}

\definecolor{summer_sky}{RGB}{58, 151, 233}
\definecolor{chateau_green}{RGB}{72, 179, 96}
\definecolor{matisse}{RGB}{25, 104, 167}
\definecolor{allports}{RGB}{31, 106, 125}
\definecolor{sun_shade}{RGB}{255, 144, 68}
\definecolor{flamingo}{RGB}{237, 88, 85}
\definecolor{studio}{RGB}{128, 91, 160}

\definecolor{maya_blue}{RGB}{102, 204, 255}
\definecolor{feijoa}{RGB}{178,223,138}
\definecolor{sushi}{RGB}{117, 168, 47}
\definecolor{norway}{RGB}{158, 194, 132}
\definecolor{japanese_laurel}{RGB}{53, 116, 40}
\definecolor{see_green}{RGB}{161,228,195}
\definecolor{monte_carlo}{RGB}{135,204,194}
\definecolor{granny_smith_apple}{RGB}{150,214,150}
\definecolor{moss_green}{RGB}{170,216,176}
\definecolor{chateau_green}{RGB}{72, 179, 96}
\definecolor{opal}{RGB}{164,207,190}
\definecolor{acapulco}{RGB}{117, 170, 148}
\definecolor{viridian}{RGB}{55, 137, 122}
\definecolor{amazon}{RGB}{56, 123, 84}
\definecolor{asparagus}{RGB}{123, 160, 91}
\definecolor{fruit_salad}{RGB}{91, 160, 94}
\definecolor{puerto_rico}{RGB}{72, 179, 150}
\definecolor{mountain_meadow}{RGB}{0, 163, 136}
\definecolor{matisse}{RGB}{25, 104, 167}
\definecolor{allports}{RGB}{31, 106, 125}
\definecolor{astral}{RGB}{55, 111, 137}
\definecolor{spring_leaves}{RGB}{46, 83, 117}
\definecolor{biscay}{RGB}{44, 62, 80}
\definecolor{midnight}{RGB}{0, 29, 50}
\definecolor{amethyst}{RGB}{153, 102, 204}
\definecolor{studio}{RGB}{128, 91, 160}
\definecolor{tapestry}{RGB}{194, 109, 132}
\definecolor{atomic_tangerine}{RGB}{255, 153, 102}
\definecolor{amber}{RGB}{255, 191, 0}
\definecolor{casablanca}{RGB}{244, 178, 84}
\definecolor{california}{RGB}{233, 140, 58}
\definecolor{tomato}{RGB}{255, 97, 56} 
\definecolor{alizarin}{RGB}{233, 58, 64}

\definecolor{linen}{RGB}{251, 239, 227}
\definecolor{double_pearl_lusta}{RGB}{253, 242, 208}
\definecolor{oasis}{RGB}{253, 242, 208}
\definecolor{milan}{RGB}{255, 254, 169}
\definecolor{texas}{RGB}{245, 232, 123}
\definecolor{maize}{RGB}{249, 212, 156}

\definecolor{turmeric}{RGB}{211, 178, 76}
\definecolor{saffron}{RGB}{249,193,62}
\definecolor{my_sin}{RGB}{255, 176, 59}
\definecolor{tree_poppy}{RGB}{246, 154, 27}
\definecolor{jaffa}{RGB}{240, 131, 58}
\definecolor{crusta}{RGB}{254, 127, 44}
\definecolor{tahiti_gold}{RGB}{223, 102, 36}
\definecolor{outrageous_orange}{RGB}{255, 100, 45}
\definecolor{safety_orange}{RGB}{254, 106, 0}

\definecolor{azalea}{RGB}{251, 196, 196}
\definecolor{oyster_pink}{RGB}{238,206,205} 
\definecolor{coral_candy}{RGB}{242,208,205} 
\definecolor{baby_pink}{RGB}{246, 194, 192}
\definecolor{petite_orchid}{RGB}{223, 157, 155}
\definecolor{apricot}{RGB}{241,140,122}
\definecolor{NY_pink}{RGB}{228,136,113}
\definecolor{carmine_pink}{RGB}{231, 76, 60}
\definecolor{deep_carmine_pink}{RGB}{236, 50, 67}

\definecolor{wewak}{RGB}{244, 143, 150}
\definecolor{light_coral}{RGB}{244, 127, 123}
\definecolor{bittersweet}{RGB}{255,111,105}
\definecolor{carnation}{RGB}{245, 80, 86}
\definecolor{flamingo}{RGB}{237, 88, 85}
\definecolor{sunset_orange}{RGB}{242,89,75}
\definecolor{ku_crimson}{RGB}{243, 0, 25}
\definecolor{amaranth}{RGB}{234,46,73}
\definecolor{valencia}{RGB}{214, 87, 70}
\definecolor{chilean_fire}{RGB}{215, 87, 44}
\definecolor{mexican_red}{RGB}{170, 41, 37}

\definecolor{napa}{RGB}{163, 154, 137}

\definecolor{athens_gray}{RGB}{236, 240, 241}
\definecolor{gallery}{RGB}{240,240,240}
\definecolor{mercury}{RGB}{230,230,230}
\definecolor{platinum}{RGB}{228,228,228}
\definecolor{silver}{RGB}{191,191,191}
\definecolor{aluminum}{RGB}{153,153,153}
\definecolor{ship_gray}{RGB}{77,77,77}
\definecolor{tuatara}{RGB}{67, 67, 67}

\definecolor{malibu}{RGB}{110, 180, 240}
\definecolor{celestial_blue}{RGB}{52, 152, 219}
\definecolor{curious_blue}{RGB}{41, 128, 185}
\definecolor{french_blue}{RGB}{0, 112, 182}
\definecolor{matisse}{RGB}{25, 104, 167}
\definecolor{shakespeare}{RGB}{85, 154, 193}
\definecolor{seagull}{RGB}{128,177,211}
\definecolor{jelly_bean}{RGB}{45, 126, 150}
\definecolor{venice_blue}{RGB}{87, 135, 105}
\definecolor{boston_blue}{RGB}{68, 147, 161}

\definecolor{turquoise}{RGB}{41,217,194}
\definecolor{java}{RGB}{2,190,196}
\definecolor{riptide}{RGB}{141,211,199}
\definecolor{mountain_meadow}{RGB}{0, 163, 136}
\definecolor{free_speech_aquamarine}{RGB}{0, 156, 114}

\definecolor{cosmic_latte}{RGB}{222, 247, 229}
\definecolor{chinook}{RGB}{163, 232, 178}
\definecolor{padua}{RGB}{121, 189, 143}
\definecolor{ocean_green}{RGB}{79, 176, 112}
\definecolor{pastel_green}{RGB}{107, 227, 135}
\definecolor{chateau_green}{RGB}{69, 191, 85}
\definecolor{RoyalBlue}{RGB}{69, 191, 85}
\definecolor{pigment_green}{RGB}{0, 175, 79}
\definecolor{fern}{RGB}{101,197,117}
\definecolor{killarney}{RGB}{56, 113, 66}

\newcommand{\unvicon}{\scalebox{0.8}{\tiny\faIcon{university}}}
\newcommand{\shieldicon}{\scalebox{0.8}{\tiny\faIcon{shield-alt}}}

\title{\textit{BaseCal}: Unsupervised Confidence Calibration via Base Model Signals}

\author{Hexiang Tan$^{\shieldicon\unvicon}$  \hspace{0.6em}  Wanli Yang$^{\shieldicon\unvicon}$ \hspace{0.6em} \textbf{Junwei Zhang}$^{\unvicon}$ \hspace{0.6em}  \textbf{Xin Chen}  \hspace{0.5em}  \textbf{Rui Tang}$^{\unvicon}$ \\ 
   \textbf{Du Su}$^{\shieldicon}$  \hspace{0.5em}  \textbf{Jingang Wang} \hspace{0.5em}  \textbf{Yuanzhuo Wang}$^{\shieldicon}$  \hspace{0.5em} \textbf{Fei Sun}\textsuperscript{\shieldicon\,\tiny\textcolor{matisse}{\faIcon[regular]{envelope}}} \hspace{0.5em} \textbf{Xueqi Cheng}$^{\shieldicon\unvicon}$\\
  $^{\shieldicon}$State Key Laboratory of AI Safety, Institute of Computing Technology, CAS\\
  $^{\unvicon}$University of Chinese Academy of Sciences \hspace{2.5em}\\
 \tt{tanhexiang21s@ict.ac.cn \,\,\, chenxin061@gmail.com \,\,\, \textsuperscript{\tiny\textcolor{matisse}{\faIcon[regular]{envelope}}}sunfei@ict.ac.cn}}

\begin{document}
\maketitle

\renewcommand*{\thefootnote}{\tiny\textcolor{matisse}{\faIcon[regular]{envelope}}}
\footnotetext{Corresponding author: Fei Sun (\href{sunfei@ict.ac.cn}{sunfei@ict.ac.cn})}
\renewcommand*{\thefootnote}{\arabic{footnote}}

\begin{abstract}
Reliable confidence is essential for trusting the outputs of LLMs, yet widely deployed post-trained LLMs  (PoLLMs) typically compromise this trust with severe overconfidence.
In contrast, we observe that their corresponding base LLMs often remain well-calibrated.
This naturally motivates us to calibrate PoLLM confidence using the base LLM as a reference.
This work proposes two ways to achieve this.
A straightforward solution, \textbf{BaseCal-ReEval}, evaluates PoLLM's responses by feeding them into the base LLM to get average probabilities as confidence.
While effective, this approach introduces additional inference overhead.
To address this, we propose \textbf{BaseCal-Proj}, which trains a lightweight projection to map the final-layer hidden states of PoLLMs back to those of their base LLMs.
These projected states are then processed by the base LLM's output layer to derive base-calibrated confidence for PoLLM's responses.
Notably, BaseCal is an unsupervised, plug-and-play solution that operates without human labels or LLM modifications.
Experiments across five datasets and three LLM families demonstrate the effectiveness of BaseCal, reducing Expected Calibration Error (ECE) by an average of \textbf{42.90\%} compared to the best unsupervised baselines\footnote{Code: \href{https://github.com/Tan-Hexiang/BaseCal}{https://github.com/Tan-Hexiang/BaseCal}}.

\end{abstract}

\section{Introduction}

Hallucinations have become a critical challenge for large language models (LLMs). 
To mitigate this risk, a primary direction is to equip model responses with confidence scores that are well-calibrated with their actual accuracy \citep{on_the_calibration, calibration_survey}. 
Such confidence enables abstention from low-confidence answers to mitigate hallucinations or otherwise alert users to potential errors.
However, widely adopted post-trained language models (PoLLMs) have been found to exhibit significant overconfidence, often assigning high confidence even to incorrect responses \citep{achiam2023gpt,zhu-etal-2023-calibration}.

\begin{figure}[t]
    \centering
    \includegraphics[width=\linewidth]{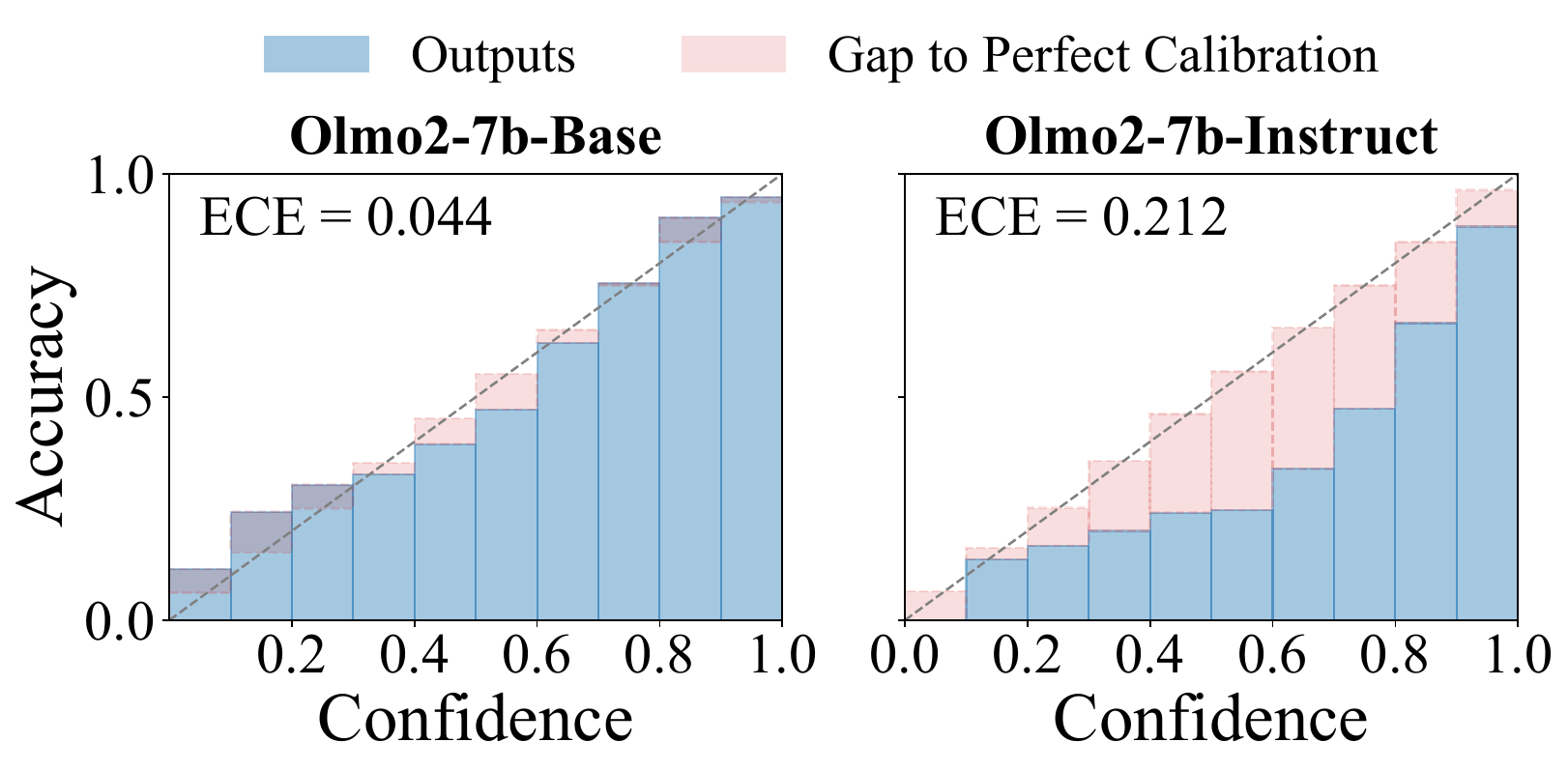}
    \captionsetup{skip=3pt}
    \caption{Calibration plots of base LLM (left) and PoLLM (right) on TriviaQA. The dashed line indicates perfect calibration; bars below it denote overconfidence.}
    \label{intro:calibration}
\end{figure}

Considerable efforts have been devoted to mitigating this overconfidence.
One direction involves supervised strategies, such as calibration-oriented fine-tuning \citep{wang-etal-2025-towards-objective, xiao2025restoring} or temperature scaling using human annotation \citep{on_the_calibration}.
However, these supervised methods rely heavily on human-labeled data, which is often difficult to obtain in real-world applications.
In contrast, unsupervised approaches attempt to estimate confidence from PoLLM itself, such as aggregated token probabilities \citep{malinin2021uncertainty}, verbalized confidence \citep{tian-etal-2023-just, xiong2024can}, and sampling-based consistency \citep{farquhar2024detecting, semantic_entropy}. 
While alleviating the reliance on human labels, these methods remain constrained by the quality of PoLLM signals, which often retain a certain degree of miscalibration \citep{tan-etal-2025-consistent, simhi-etal-2025-trust}.

Inspired by findings that base LLMs are well-calibrated in multiple-choice tasks \citep{luo2025your, xiao2025restoring}, we investigate whether they can serve as an external reference to enhance PoLLM calibration in free-form factual QA tasks.
Figure \ref{intro:calibration} (more results in \S \ref{sec:motivation}) illustrates that base LLMs exhibit significantly superior confidence calibration compared to PoLLMs on the widely used QA dataset TriviaQA.
Motivated by this observation, we seek to calibrate PoLLMs with their base counterparts.

A straightforward solution, \textbf{BaseCal-ReEval}, is to feed the PoLLM's response into its corresponding base LLM and utilize the base LLM's average token probabilities for this response as the confidence.
While simple and effective, BaseCal-ReEval necessitates an additional forward pass on the base LLM, which significantly increases inference latency and resource consumption.

To mitigate this cost, we propose \textbf{BaseCal-Proj}, which approximates the base LLM's confidence via a lightweight projection.
Specifically, we extract the final-layer states of PoLLM and base LLM for the same input questions and PoLLM-generated responses, and train a projection module to map the state of PoLLM to the target base state.
During inference, the PoLLM's final-layer states are projected and passed through the base LLM's output layer to approximate the base probability distribution.
We then calculate the confidence by averaging the probabilities assigned to the PoLLM-generated tokens under this distribution.
Since BaseCal-Proj involves only a lightweight projection and an output layer, it incurs negligible overhead compared to BaseCal-ReEval and sampling-based methods \citep{semantic_entropy,farquhar2024detecting}.

Experiments across five datasets and three LLM families demonstrate the significant effectiveness of our methods. 
Compared to the best unsupervised baselines, BaseCal-ReEval reduces ECE by an average of 42.90\%, and BaseCal-Proj achieves a 35.32\% reduction but with significantly lower inference overhead.
Further analysis shows that (i) these improvements persist across various model scales and post-training strategies; and (ii) BaseCal-Proj exhibits strong generalization capabilities on unseen questions.
In conclusion, BaseCal serves as a plug-and-play framework to restore PoLLM calibration without parameter modification, making it easily adaptable to real-world applications.

\section{Related Work}

\subsection{Overconfidence of LLMs}

Emerging evidence indicates that post-training can induce systematic overconfidence, thus degrading model calibration \citep{xiao2025restoring, leng2025taming, wang-etal-2025-towards-objective}.
This effect occurs in both instruction tuning and RLHF \citep{zhu-etal-2023-calibration, leng2025taming}, and strengthens with more tunable parameters \citep{chen-etal-2023-close}.
Prior work attributes this phenomenon to factors such as data overlap between fine-tuning and pretraining corpora \citep{wang-etal-2025-towards-objective}, reward bias \citep{leng2025taming}, preference collapse \citep{xiao2025restoring}, and catastrophic forgetting \citep{he2023preserving}.
To mitigate overconfidence, prior work has explored domain-specific fine-tuning \citep{xiao2025restoring}, confidence-aware reward \citep{leng2025taming}, and feature-preserving adaptation \citep{he2023preserving}.

\subsection{Confidence Calibration for LLMs}

Confidence calibration aims to align model confidence with the correctness of its predictions.
On the one hand, \textbf{supervised methods} rely on human-labeled data to perform calibration, e.g., temperature scaling \citep{on_the_calibration, xie2024calibrating, joy2023sample, yu2022robust} optimizes a single temperature parameter to rescale probabilities.
Other methods fine-tune models to provide calibrated confidence \citep{kapoor-etal-2024-calibration, tao-etal-2024-trust, chen-etal-2023-close}.
Despite their effectiveness, supervised methods suffer from both reliance on human-labeled data \citep{shen2024thermometer} and limited generalization \citep{liu2025on}.

In contrast, \textbf{unsupervised methods} avoid reliance on human-annotated labels.
\textit{Prompt-based methods} directly query the model about the correctness of its own output, including P(true), which estimates confidence as the probability that the model judges its answer to be correct \citep{kadavath2022languagemodelsmostlyknow}, and verbalized confidence, where confidence is explicitly expressed in natural language \citep{tian-etal-2023-just, zhang-etal-2024-calibrating, lin2022teaching, xiong2024can}.
\textit{Sampling-based methods} estimate confidence from the uncertainty across response samples, either via semantic agreement \citep{manakul-etal-2023-selfcheckgpt, xiong2024can, lin2024generating, raj2023semantic} or through semantic entropy \citep{farquhar2024detecting, semantic_entropy, nikitin2024kernel}.
However, these methods remain limited by their reliance on signals from the PoLLMs themselves, which preserve some extent of overconfidence \citep{tan-etal-2025-consistent, simhi-etal-2025-trust}.

\begin{figure*}[t]
    \centering
    \includegraphics[width=\linewidth]{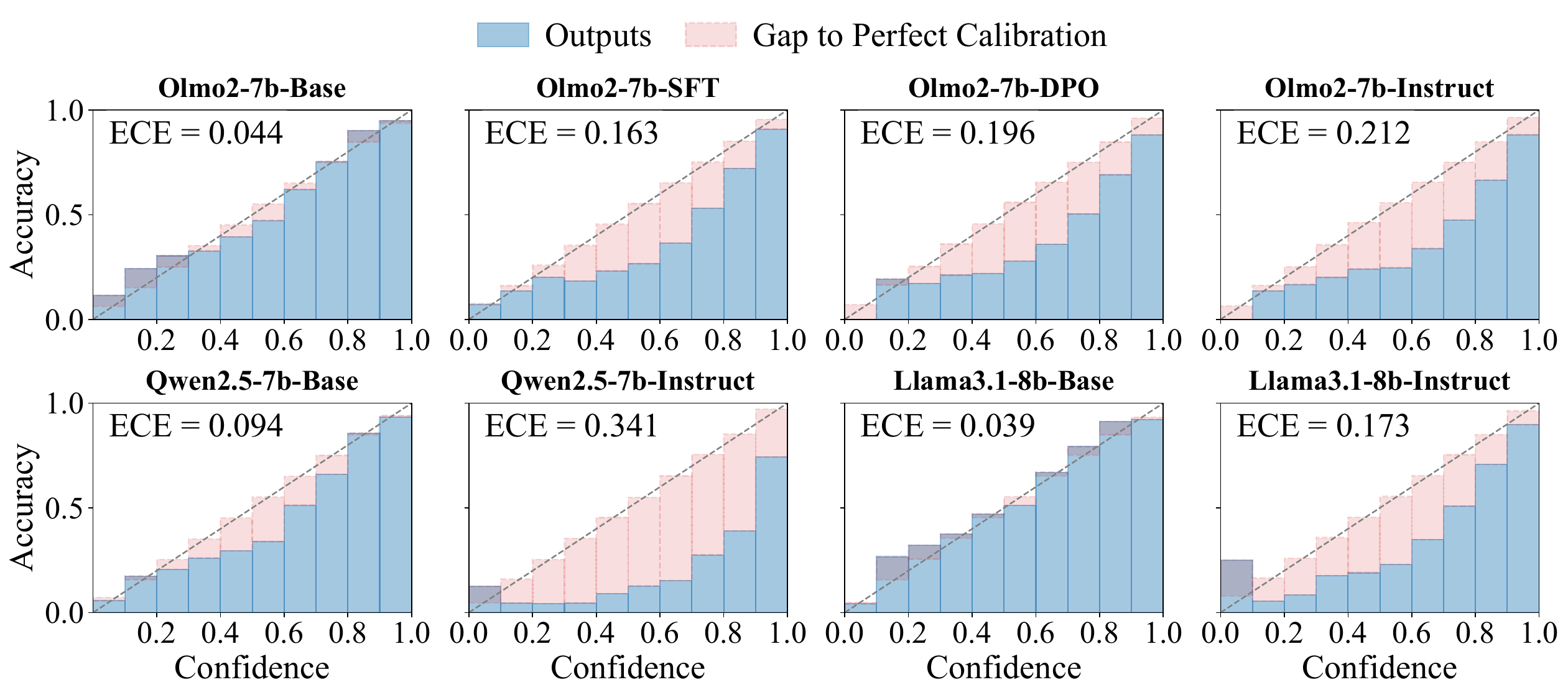}
    \caption{Calibration plots on TriviaQA across three model families. The top row presents Olmo2 checkpoints after different post-training stages. The dashed line is perfect calibration, while bars below denote overconfidence.}
    \label{fig:calibration_instruct_vs_base}
\end{figure*}

Closely related to our study, \citet{luo2025your} optimizes a temperature parameter to rescale PoLLM's output probabilities to match those of base LLMs. 
In contrast to their probability-level adjustments, we recover calibration from the model’s hidden states, thereby leveraging richer internal information. 
More importantly, while their method is restricted to multiple-choice formats, our approach is more general and natively supports free-form generation, which is the predominant interaction paradigm for current LLMs.

\begin{figure*}[!t]
    \centering
    \begin{minipage}[b]{0.49\linewidth}
        \centering
        \includegraphics[width=\linewidth]{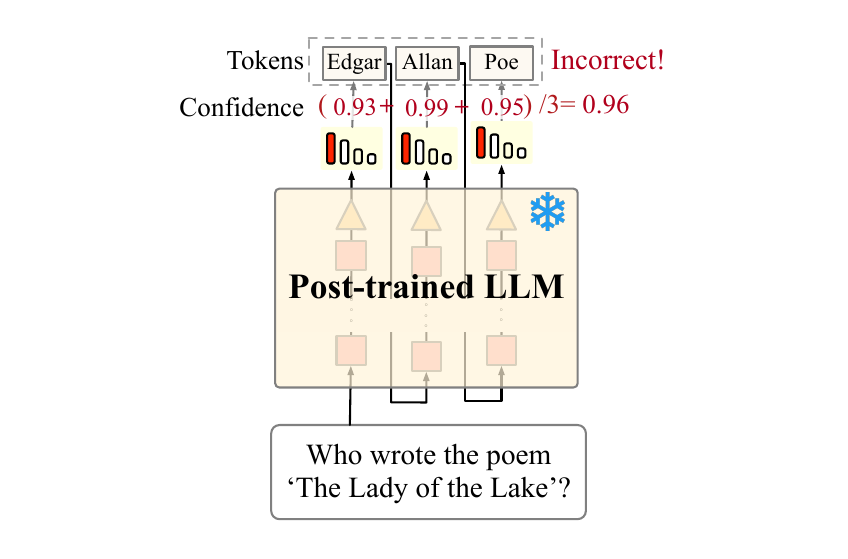}
        \subcaption{Vanilla}
        \label{fig:vanilla}
    \end{minipage}\hfill
    \begin{minipage}[b]{0.49\linewidth}
        \centering
        \includegraphics[width=\linewidth]{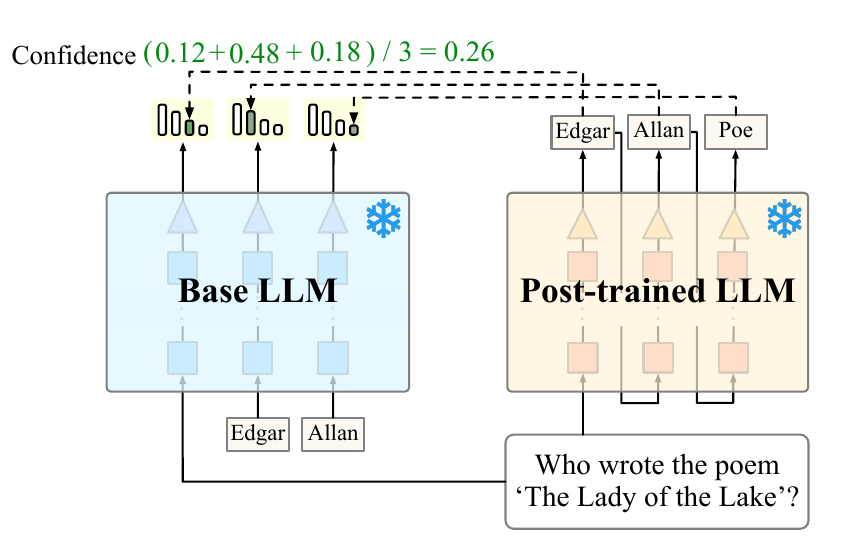}
        \subcaption{BaseCal-ReEval}
        \label{fig:BaseCal-ReEval}
    \end{minipage}\hfill
    \\
    \begin{minipage}[b]{0.49\linewidth}
        \centering
        \includegraphics[width=\linewidth]{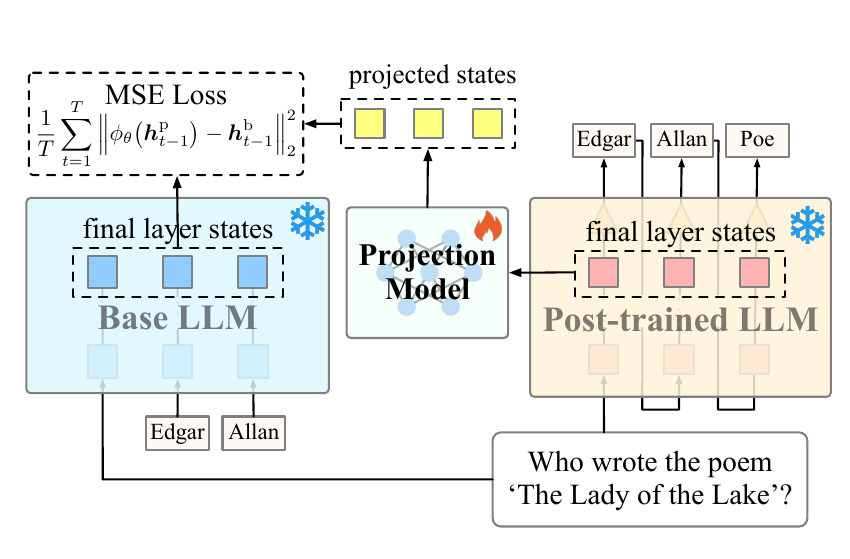}
        \subcaption{BaseCal-Proj Training}
        \label{fig:BaseCal-Proj-Training}
    \end{minipage}\hfill
    \begin{minipage}[b]{0.49\linewidth}
        \centering
        \includegraphics[width=\linewidth]{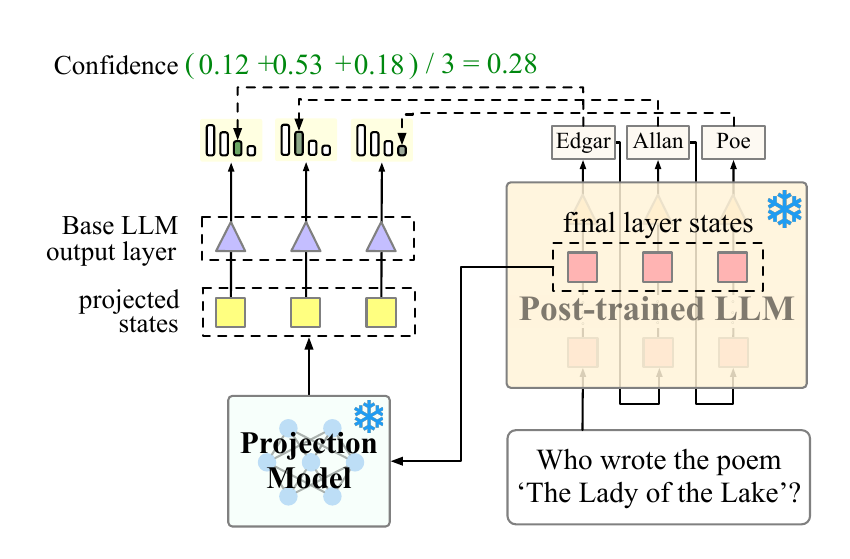}
        \subcaption{BaseCal-Proj Inference}
        \label{fig:BaseCal-Proj}
    \end{minipage}

    \caption{Frameworks of three methods. \textbf{(a) Vanilla:} use the PoLLM’s own aggregated probabilities as confidence. \textbf{(b) BaseCal-ReEval:} evaluate the PoLLM-generated response with the corresponding base LLM. 
    \textbf{(c) BaseCal-Proj Training:} train a lightweight projection from PoLLM to base LLM final-layer hidden states. \textbf{(d) BaseCal-Proj Inference:} use the learned projection and the base output layer to estimate confidence.}

    \label{fig:compare_methods}
\end{figure*}

\section{Preliminaries}
\label{sec:Preliminaries}

Confidence calibration aims to align model confidence with the actual correctness probability of LLMs' responses.
Let $\mathcal{D} = \{(\boldsymbol{x}_i, \boldsymbol{y}_i^{*})\}, i=1, \dots, N$ denote a dataset of $N$ samples, where $\boldsymbol{x}_i$ is an input prompt and $\boldsymbol{y}^{*}_i$ is the corresponding ground-truth answer. 
Given $\boldsymbol{x}_i$, the model $\mathcal{M}$ generates a response $\boldsymbol{y}_i = (y_1^{(i)},\dots, y_T^{(i)})$, where $T$ is the length of the response.
The response correctness $z_i \in \{0,1\}$ is commonly evaluated by comparing the response with the ground-truth $\boldsymbol{y}^{*}_i$.
For notational simplicity, we omit the subscript $i$ when the context is unambiguous.
A model is perfectly calibrated if
\begin{equation}
P(z = 1 \mid \text{c}(\boldsymbol{x},\boldsymbol{y}) = q) = q.
\end{equation}
where $\text{c}(\boldsymbol{x}, \boldsymbol{y}) \in [0, 1]$ is the confidence function.
This means that among all responses assigned a confidence of $q$ (e.g., 70\%), the proportion of correct answers should be $q$.  

To quantify the degree of miscalibration, we employ the Expected Calibration Error (\textbf{ECE}) \citep{on_the_calibration}, which partitions the samples into $M$ equal-width confidence bins $\{B_1, \dots, B_M\}$, then calculates the absolute gap between average accuracy $\text{acc}(B_m)$ and average confidence $\text{conf}(B_m)$:
\begin{equation}
\text{ECE} = \sum_{m=1}^M \frac{|B_m|}{N} \left| \text{acc}(B_m) - \text{conf}(B_m) \right|.
\end{equation}

We primarily focus on free-form QA as it aligns better with the primary usage paradigm of current LLMs.
In this task, the aggregated token probability \citep{malinin2021uncertainty} is commonly treated as the \textbf{Vanilla} confidence, as it represents the most direct utilization of the model's signal, without requiring specific prompt design \citep{tian-etal-2023-just} or multiple sampling \citep{farquhar2024detecting}.
We employ average aggregation following \citet{orgad2024llmsknowshowintrinsic} to mitigate the effect of sequence length.
\begin{equation}
\text{c}(\boldsymbol{x},\boldsymbol{y})=\frac{1}{T}\sum_{t=1}^{T} P(y_t \mid \boldsymbol{x},\boldsymbol{y}_{<t}).
\end{equation}

\section{Calibration with Base LLMs}

\subsection{Motivation}
\label{sec:motivation}

Current unsupervised calibration methods are limited by the signals from PoLLMs, e.g., verbalized confidences are often very high \citep{leng2025taming}, and consistency can fail under self-consistent errors \citep{tan-etal-2025-consistent}.
This motivates us to identify a more reliable external reference beyond PoLLMs.
Inspired by findings that base LLMs are well-calibrated in multiple-choice tasks \citep{luo2025your, xiao2025restoring, achiam2023gpt}, we investigate whether this superior calibration persists in more challenging free-form QA tasks.

The experiments are conducted on the widely used QA dataset TriviaQA, using Qwen2.5 \citep{yang2024qwen2}, Llama3.1 \citep{grattafiori2024llama3}, and the fully open-source Olmo2 \citep{olmo20242}, which provide checkpoints at different post-training stages.
This enables us to analyze the effects of various post-training strategies.
As shown in Figure \ref{fig:calibration_instruct_vs_base}, PoLLMs are consistently more miscalibrated than their base counterparts across all three model families, indicating that the better calibration of base LLMs persists in free-form QA. 
Moreover, results on Olmo2 indicate that diverse post-training methods all impair confidence calibration, revealing a shared limitation of existing post-training approaches.
Leveraging these insights, our core idea is to calibrate PoLLMs with their base counterparts, thereby eliminating the reliance on miscalibrated signals from the PoLLMs.

\subsection{BaseCal-ReEval}
As a direct instantiation of our idea, BaseCal-ReEval evaluates the PoLLM's generation by passing the identical input sequence through the base LLM to extract the corresponding target token probabilities as the confidence score (Figure \ref{fig:BaseCal-ReEval}).

Let $\mathcal{M}_\text{p}$ and $\mathcal{M}_\text{b}$ denote the PoLLM and its corresponding base LLM, respectively. 
Given an input prompt $\boldsymbol{x}$, let $\boldsymbol{y}^\text{p} = (y_1^\text{p}, \dots, y_T^\text{p})$ denote the response sequence generated by $\mathcal{M}_\text{p}$.
To quantify the base LLM's confidence in $\boldsymbol{y}^\text{p}$, we employ $\mathcal{M}_\text{b}$ to score the sequence.
Specifically, for each position $t$, we compute the probability that $\mathcal{M}_\text{b}$ assigns to the token $y_t^\text{p}$ produced by $\mathcal{M}_\text{p}$, conditioned on the prompt and the preceding tokens $\boldsymbol{y}_{<t}^\text{p}$.
The sequence-level confidence $c_{\text{b}}(\boldsymbol{x}, \boldsymbol{y})$ is then defined as the average probability of the generated tokens under the base LLM:
\begin{equation}\text{c}_{\text{b}}(\boldsymbol{x}, \boldsymbol{y}^\text{p}) = \frac{1}{T} \sum_{t=1}^{T} P_{\mathcal{M}_\text{b}}(y_t^\text{p} \mid \boldsymbol{x}, \boldsymbol{y}_{<t}^\text{p}).\label{eq:seq_conf}\end{equation}

\subsection{BaseCal-Proj}
Despite its effectiveness, BaseCal-ReEval necessitates a full forward pass of the base LLM, incurring additional computational and memory overhead during inference. 
To mitigate this, we propose BaseCal-Proj, which learns a lightweight projection to map the PoLLM's final-layer hidden states into the base LLM's representation space. 
These projected states are then fed into the base LLM's output layer to derive the final probability, thereby bypassing the transformer blocks while preserving the base LLM's calibration benefit.

\paragraph{Prepare Hidden State Pairs.}
To learn the mapping from PoLLM hidden states to those of the base LLM, we extract the final-layer hidden states from both models, conditioned on \textit{identical contexts}.
Concretely, for the $t$-th token $y_t^\text{p}$, we extract the final-layer hidden state from the PoLLM, conditioned on the prompt and the prefix response:
\begin{equation}
\boldsymbol{h}_{t-1}^{\text{p}}
=
\mathcal{M}_{\text{p}}^{(L)}\!\left(\boldsymbol{x};\boldsymbol{y}_{<t}^\text{p}\right),
\label{eq:polm_hidden}
\end{equation}
where $\mathcal{M}^{(L)}(\cdot)$ denotes the mapping from the input sequence to the $L$-th (final) layer hidden state.
Similarly, we feed the same concatenated sequence $(\boldsymbol{x},\boldsymbol{y}_{<t}^\text{p})$ into the base LLM to obtain the final-layer hidden state:
\begin{equation}
\boldsymbol{h}_{t-1}^{\text{b}}
=
\mathcal{M}_{\text{b}}^{(L)}\!\left(\boldsymbol{x};\boldsymbol{y}_{<t}^\text{p}\right).
\label{eq:base_hidden}
\end{equation}
Repeating this procedure for all positions $t=1,\dots,T$, we get a sequence of hidden-state pairs
$\{(\boldsymbol{h}^{\text{p}}_{0},\boldsymbol{h}^{\text{b}}_{0}),\dots,(\boldsymbol{h}^{\text{p}}_{T-1},\boldsymbol{h}^{\text{b}}_{T-1})\}$ for each question.

\paragraph{Projection Training.}
Our goal is to minimize the discrepancy between the projected PoLLM hidden states and corresponding base-model representations. 
This objective requires two components: a projection model $\phi_{\theta}$ and a loss function $\mathcal{L}$.
For our primary implementation, we adopt a one-layer linear projection for $\phi_{\theta}$ and Mean Squared Error (MSE) for $\mathcal{L}$ as a simple yet effective instantiation: 
\begin{equation}
\begin{aligned}
\mathcal{L}(\theta)
&=
\frac{1}{T}\sum_{t=1}^{T}
\left\|
\phi_{\theta}\!\left(\boldsymbol{h}^{\text{p}}_{t-1}\right)
-
\boldsymbol{h}^{\text{b}}_{t-1}
\right\|_2^2, \\
\phi_{\theta}\!\left(\boldsymbol{h}^{\text{p}}_{t-1}\right)
&=
\boldsymbol{W}\boldsymbol{h}^{\text{p}}_{t-1} + \boldsymbol{b},
\end{aligned}
\label{eq:mse_proj}
\end{equation}
where $\boldsymbol{W}\in\mathbb{R}^{d\times d}$ and $\boldsymbol{b}\in\mathbb{R}^{d}$ are learnable parameters.
During training, we freeze both $\mathcal{M}_{\text{p}}$ and $\mathcal{M}_{\text{b}}$, optimizing only the projection parameters $\theta$.
Furthermore, we explore non-linear architectures and alternative loss functions in \Cref{analyse:projection_model} and Appendix \ref{appendix:loss}, respectively.

\paragraph{Inference.}
Figure \ref{fig:BaseCal-Proj} shows the pipeline of BaseCal-Proj at inference time.
We extract the final-layer hidden states of the post-trained LLM $\mathcal{M}_{\text{p}}$, and transform them via the learned projection $\phi_{\theta}$. 
These projected states are then fed into the base LLM output layer $\boldsymbol{W}^{\text{o}}_{\text{b}}$ to get the probabilities of \textit{target tokens} $[y_1^\text{p}, \dots, y_T^\text{p}]$, which are then aggregated to the sequence-level confidence:
\begin{equation}
\tilde{\text{c}}(\boldsymbol{x},\boldsymbol{y}^{\text{p}})\! = \! \frac{1}{T}\!\sum_{t=1}^{T}\!\text{softmax}(\boldsymbol{W}^{\text{o}}_\text{b} \, \phi_{\theta}\!\left(\boldsymbol{h}^{\text{p}}_{t-1}\right))[y_t^\text{p}].
\end{equation}
Notably, the projected states are used only for confidence estimation. 
The PoLLM still performs generation with its original hidden states, thereby preserving the generation quality of PoLLM. 
Moreover, compared to standard inference, BaseCal-Proj introduces only a lightweight projection model and the base-model output layer, resulting in nearly negligible inference-time overhead.

\begin{table*}[t]
  \centering
  \begin{adjustbox}{max width=\linewidth}
  \begin{tabular}{ll *{12}{c}}
    \toprule
    \multirow{2.5}{*}{\textbf{Model}} &
    \multirow{2.5}{*}{\textbf{Method}} &
    \multirow{2.5}{*}{\textbf{Unsupervised}} &
    \multicolumn{2}{c}{\textbf{TriviaQA}} &
    \multicolumn{2}{c}{\textbf{NQ}} &
    \multicolumn{2}{c}{\textbf{WebQ}} &
    \multicolumn{2}{c}{\textbf{SQuAD}} &
    \multicolumn{2}{c}{\textbf{MMLU}} \\
    \cmidrule(lr){4-5} \cmidrule(lr){6-7} \cmidrule(lr){8-9} \cmidrule(lr){10-11} \cmidrule(lr){12-13}
    & & & ECE ($\downarrow$) & BS ($\downarrow$) & ECE ($\downarrow$) & BS ($\downarrow$) & ECE ($\downarrow$) & BS ($\downarrow$) & ECE ($\downarrow$) & BS ($\downarrow$) & ECE ($\downarrow$) & BS ($\downarrow$)\\
    \midrule

    \multirow{8}{*}{\textbf{\makecell[l]{Llama3.1-8B\\-Instruct}}}
      & Temp. Scaling & \ding{55}
                           & 0.0226 & 0.1702
                           & 0.0460 & 0.1743
                           & 0.0930 & 0.2446
                           & 0.0911 & 0.1818
                           & 0.0307 & 0.2021 \\
        \cmidrule(lr){2-13}
      & Vanilla    & \ding{51}        & 0.1725 & 0.2090 & 0.4532	& 0.3882 & 0.3832 & 0.3800 & 0.5255 & 0.4546  & 0.1071 & 0.2152 \\
      & P(true)     & \ding{51}        & 0.2476 & 0.2506 & 0.4439	& 0.3972 & 0.4581 & 0.4496 & 0.5532 & 0.5094  & 0.2971 & 0.3079 \\
      & Verbalization  & \ding{51}     & 0.1769 & 0.2046 & 0.2689 & 0.2507 & 0.3565 & 0.3456 & 0.3603 & 0.3208  & 0.2011 & 0.2633 \\
      & Semantic Entropy & \ding{51}   & 0.2443 & 0.2302 & 0.4927	& 0.4371 & 0.3635 & 0.3446 & 0.4645 & 0.4003  & 0.4085 & 0.3991 \\
      & DACA       & \ding{51}         & - & - & - & - & - & - & - & - & 0.0473 & 0.1804 \\
      \cmidrule(lr){2-13}
      &  BaseCal-Proj & \ding{51} 
                           &  \underline{0.0387} &  \underline{0.1850}
                           &  \underline{0.2488} &  \underline{0.2358}
                           &  \underline{0.2091} &  \underline{0.2764}
                           &  \underline{0.3134} &  \underline{0.2816}
                           &  \textbf{0.0336} &  \underline{0.1500} \\
      &  BaseCal-ReEval & \ding{51} 
                           &  \textbf{0.0309} &  \textbf{0.1724}
                           &  \textbf{0.2462} &  \textbf{0.2349}
                           &  \textbf{0.1873} &  \textbf{0.2737}
                           &  \textbf{0.2959} &  \textbf{0.2755}
                           &  \underline{0.0375} &  \textbf{0.1473} \\
    \midrule
    \multirow{8}{*}{\textbf{\makecell[l]{Qwen2.5-7B\\-Instruct}}}
      & Temp. Scaling & \ding{55} 
                           & 0.0895 & 0.1850
                           & 0.1304 & 0.1938
                           & 0.0738 & 0.2341
                           & 0.0966 & 0.2007
                           & 0.2261 & 0.2354 \\
        \cmidrule(lr){2-13}
      & Vanilla     & \ding{51}       & 0.3406 & 0.3118 & 0.5562	& 0.4839 & 0.4486 & 0.4328 & 0.6012 & 0.5550  & 0.2569 & 0.2607 \\
      & P(true)      & \ding{51}       & 0.2113 & 0.2134 & 0.3723	& 0.3708 & 0.4838 & 0.4829 & 0.5666 & 0.5650 & 0.3204 & 0.3256 \\
      & Verbalization  & \ding{51}     & 0.2889 & 0.3031 & 0.4718	& 0.4623 & 0.3977 & 0.3957 & 0.4800 & 0.4646  & 0.1972 & 0.2546 \\
      & Semantic Entropy & \ding{51}   & 0.3583 & 0.3191 & 0.5192	& 0.4502 & 0.2497 & 0.2853 & \underline{0.4041} & 0.3821  & 0.2858 & 0.2856 \\
      & DACA       & \ding{51}        & - & - & - & - & - & - & - & - & \underline{0.0703} & \underline{0.1618} \\
      \cmidrule(lr){2-13}
      &  BaseCal-Proj & \ding{51} 
                           &  \underline{0.1393} &  \underline{0.1923}
                           &  \underline{0.3382} &  \underline{0.2880}
                           &  \textbf{0.1792} &  \textbf{0.2696}
                           &  0.4085 &   \underline{0.3569}
                           &  0.0889 &  0.1662 \\
      &  BaseCal-ReEval & \ding{51} 
                           &  \textbf{0.1120} &  \textbf{0.1789}
                           &  \textbf{0.3161} &  \textbf{0.2714}
                           &  \underline{0.1983} &  \underline{0.2801}
                           &  \textbf{0.3215} &  \textbf{0.2980}
                           &  \textbf{0.0393} &  \textbf{0.1465} \\
    \midrule
    \multirow{8}{*}{\textbf{\makecell[l]{Olmo2-7B\\-Instruct}}}
      & Temp. Scaling & \ding{55} 
                           & 0.0286 & 0.1939
                           & 0.0742 & 0.1826
                           & 0.0674 & 0.2191
                           & 0.0587 & 0.1769
                           & 0.1707 & 0.2275 \\
        \cmidrule(lr){2-13}
      & Vanilla    & \ding{51}         & 0.2121 & 0.2465 & 0.4404	& 0.3739 & 0.3638 & 0.3514 & 0.4459 & 0.3733  & 0.2465 & 0.2762 \\
      & P(true)      & \ding{51}       & 0.2055 & 0.2393 & 0.3120 & 0.3166 & 0.4071 & 0.4043 & 0.4822 & 0.4582  & 0.1940 & 0.2549 \\
      & Verbalization  & \ding{51}     & 0.2054 & 0.2721 & \textbf{0.2030} & \underline{0.2478} & 0.2537 & 0.3101 & 0.2956 & 0.3040  & 0.1533 & 0.2747 \\
      & Semantic Entropy  & \ding{51}  & 0.1910 & 0.2197 & 0.4178 & 0.3646 & 0.2469 & 0.2769 & 0.3906 & 0.3558  & 0.3132 & 0.3607 \\
      & DACA        & \ding{51}        & - & - & - & - & - & - & - & - & 0.0555 & 0.1898 \\
      \cmidrule(lr){2-13}
      &  BaseCal-Proj & \ding{51} 
                           &  \underline{0.0314} &  \underline{0.1966}
                           &  0.2712 &  0.2511
                           &  \underline{ 0.1967} &  \underline{0.2652}
                           &  \underline{0.2393} &  \underline{0.2357}
                           &  \underline{0.0525} &  \underline{0.1768} \\
      &  BaseCal-ReEval & \ding{51} 
                           &  \textbf{0.0269} &  \textbf{0.1947}
                           &  \underline{0.2304} &  \textbf{0.2312}
                           &  \textbf{0.1587} &  \textbf{0.2483}
                           &  \textbf{0.2131} &  \textbf{0.2216}
                           &  \textbf{0.0470} &  \textbf{0.1717} \\
    \bottomrule
  \end{tabular}
  \end{adjustbox}
  \captionsetup{skip=5pt}
  \caption{Calibration results across 5 datasets and 3 PoLLMs.
\textbf{Bold} indicates the best unsupervised calibration performance, while \underline{underlining} denotes the second best. Since DACA is specifically designed for multiple-choice tasks, it is marked with ``-'' on free-form QA datasets.}
  \label{table:main_table}
\end{table*}

\section{Experiments}
In this section, we conduct experiments to answer the following research questions:
\begin{itemize}[leftmargin=16pt] %
    \item \textbf{RQ1.} How do the proposed BaseCal-ReEval and BaseCal-Proj perform compared to the baselines in the confidence calibration task?
    \item \textbf{RQ2.} How does BaseCal-Proj generalize to unseen questions?
    \item \textbf{RQ3.} How do different projection models impact BaseCal-Proj?
    \item \textbf{RQ4.} How do the proposed methods perform across different PoLLMs, i.e., PoLLMs with different sizes and post-training strategies?
    \item \textbf{RQ5.} How does BaseCal-Proj benefit selective classification?
\end{itemize}

\subsection{Experimental Setup}

\paragraph{LLMs.} 
We conduct experiments across a diverse set of model families, including Qwen2.5 \citep{yang2024qwen2}, Llama3.1 \citep{grattafiori2024llama3}, and Olmo2 \citep{olmo20242} series. For each post-trained model, we use its corresponding pre-trained counterpart to perform calibration.

\paragraph{Datasets.} 
Our evaluation covers an extensive range of free-form question answering benchmarks, including TriviaQA \citep{triviaqa}, Natural Questions (NQ; \citealp{lee-etal-2019-latent}), SQuAD \citep{squad}, and WebQuestions (WebQ; \citealp{webquestion}).
The correctness of responses is evaluated using LLM-as-a-judge, following \citet{tian-etal-2023-just, orgad2024llmsknowshowintrinsic}. 
We further validate the effectiveness of LLM-as-a-judge via human verification.
Appendix \ref{appendix:llm-as-a-judge} shows more details.
To compare with DACA, we also include the widely-adopted multiple-choice benchmark MMLU \citep{mmlu}, which spans 57 subjects from STEM to humanities.
Appendix \ref{appendix:datasets} shows detailed statistics of these datasets.

\paragraph{Metrics.} 
Following previous works \citep{tian-etal-2023-just, kadavath2022languagemodelsmostlyknow}, we evaluate calibration using ECE \citep{on_the_calibration} with 10 bins as introduced in \Cref{sec:Preliminaries} and the Brier Score (BS; \citealp{brier}), which measures the mean squared error between confidence and correctness labels.

\paragraph{Baselines.}
Since our approach does not rely on labeled data, we primarily compare it against \textit{unsupervised} baselines. 
\begin{enumerate}[label=(\roman*),leftmargin=16pt] %
    \item \textbf{Vanilla.} This approach directly employs the model’s native probabilities as confidence. For free-form QA, we follow prior work \citep{orgad2024llmsknowshowintrinsic, factualconfidence, malinin2021uncertainty} and aggregate the token-level probabilities of the generated sequences to obtain a single confidence score.
    \item \textbf{P(True).} This method prompts the model to self-assess its own output and takes the predicted probability of answering ``True'' as the confidence score \citep{kadavath2022languagemodelsmostlyknow}.
    \item \textbf{Verbalization.} This method prompts PoLLMs to express confidence in natural language. We use the prompt from \citet{tian-etal-2023-just}.
    \item \textbf{Semantic Entropy (SE).} SE \citep{farquhar2024detecting, semantic_entropy} samples multiple responses and computes the entropy over semantic clusters derived from these samples. Following \citet{li-etal-2025-towards}, we normalize the SE values to the range $[0,1]$.
    \item \textbf{DACA.} DACA \citep{luo2025your}, which is designed specifically for the multiple-choice format, uses examples where the PoLLM and base LLM produce the same choice, and then learns a temperature to align the PoLLM's probabilities with those of the pre-trained model.
\end{enumerate}
Beyond \textit{unsupervised} baselines, we include \textbf{Temperature Scaling} (TS; \citealp{on_the_calibration}) as a \textit{supervised} reference baseline, which learns a temperature parameter by minimizing the negative log-likelihood with respect to answer correctness.
More details are shown in Appendix \ref{appendix:baselines}.

\paragraph{Implementation Details.}
The training of BaseCal-Proj leverages training set questions to derive hidden state pairs without relying on human-annotated labels. We mitigate overfitting via an early stopping mechanism, monitored by the loss on the validation set. Statistics for the training and validation sets are detailed in Appendix \ref{appendix:datasets}.

\subsection{Overall Performance Comparison (RQ1)}

To assess how well the proposed methods calibrate PoLLM confidence, we compare BaseCal-ReEval and BaseCal-Proj against all baselines.
\Cref{table:main_table} shows the results on five datasets across three PoLLMs.
We analyze the results from various perspectives and obtain the following observations:

\paragraph{(i) Comparison with Unsupervised Methods.}
As shown in \Cref{table:main_table},  BaseCal-ReEval and BaseCal-Proj reduce the average ECE by 55.65\% and 49.78\% compared to the Vanilla PoLLMs, respectively.  
Compared to other unsupervised methods, our methods achieve the best calibration performance in 29 out of 30 experimental settings (5 datasets $\times$ 3 PoLLMs $\times$ 2 metrics).
While Verbalization works well in Olmo2-7B-Instruct on NQ, its performance varies greatly and drops substantially on other models like Qwen2.5-7B.
This instability likely arises from their strong reliance on the instruction-following ability.
In contrast, BaseCal-ReEval consistently ranks among the top two across all settings, significantly outperforming Verbalization (average ECE 0.1641 vs. 0.2874).

Notably, on multiple-choice datasets, DACA exhibits performance that significantly outperforms other baselines.
However, compared to DACA, BaseCal-ReEval achieves even superior results (e.g., 0.0393 vs. 0.0703 ECE on Qwen2.5).
This underlines the efficacy of directly leveraging base LLM confidence, as opposed to rescaling probabilities to approximate the base distribution.
Meanwhile, BaseCal-Proj also remains superior or comparable to DACA, likely benefiting from the richer information within the hidden states compared to probability-level scaling by DACA.

\paragraph{(ii) Comparison with Supervised Methods.}
Despite being unsupervised, BaseCal achieves performance comparable to supervised TS on TriviaQA and MMLU.
In the remaining settings, while BaseCal underperforms TS baseline, it maintains the best performance among unsupervised methods and significantly narrows the gap with supervised approaches.
These results highlight BaseCal as a compelling, label-efficient alternative for scenarios where ground-truth data is unavailable.

\begin{figure}[t]
    \centering

    \includegraphics[width=\linewidth]{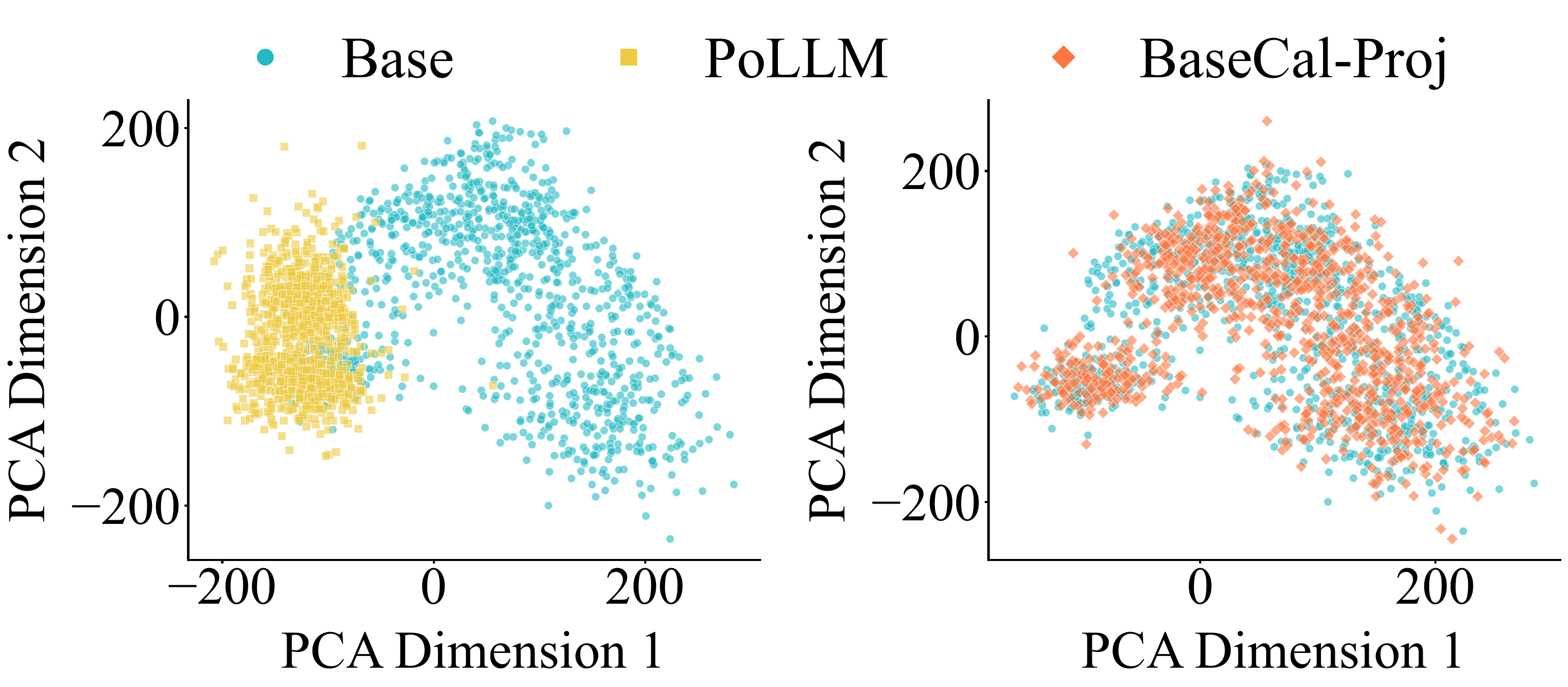} 
    \caption{Visualization of hidden states for PoLLM vs. Base LLM (left) and BaseCal-Proj vs. Base LLM (right), for 2500 randomly sampled examples on TriviaQA.}
    \label{fig:visualization_hd}
\end{figure}

\paragraph{(iii) Comparison between BaseCal-Proj and BaseCal-ReEval.}
As shown in \Cref{table:main_table}, BaseCal-Proj achieves an average ECE of 0.1859 across all settings, closely approximating the performance of BaseCal-ReEval (0.1641), which directly uses base LLMs to get confidence.
This means that BaseCal-Proj successfully recovers the calibration of base LLMs without the cost of an extra forward pass on them.
To further understand the mechanism behind this, we visualize the hidden states of PoLLM, BaseCal-Proj, and base LLM in Figure \ref{fig:visualization_hd}.
As illustrated, while the hidden states of the PoLLM are distinct from those of the base LLM, the projected states closely align with those of the base LLM.
This visualization confirms that the projection layer effectively maps the post-trained hidden states back to the base LLM, thereby explaining their comparable calibration performance.

\subsection{Generalization of BaseCal-Proj (RQ2)}
Although BaseCal-Proj does not rely on human labels, it still requires questions for training, raising the question of whether it can generalize to unseen questions. 
To investigate this, we conduct a cross-dataset generalization evaluation across four QA datasets: SQuAD, NQ, TriviaQA, and WebQ.
We compare the in-domain (ID) and out-of-domain (OOD) performance by calculating the difference in their ECE values, defined as $\Delta \text{ECE} = \text{ECE}_{\text{ID}} - \text{ECE}_{\text{OOD}}$. 
A higher $\Delta \text{ECE}$ indicates superior generalization capability, where $\Delta \text{ECE}>0$ signifies that the model’s OOD calibration outperforms its ID performance.
We compare the generalization of BaseCal-Proj against the supervised method, temperature scaling. 
Notably, BaseCal-Proj is trained solely using the questions from the training dataset, while temperature scaling utilizes both the questions and human-labeled ground truth.

\begin{figure}[t]
    \centering
    \includegraphics[width=\linewidth]{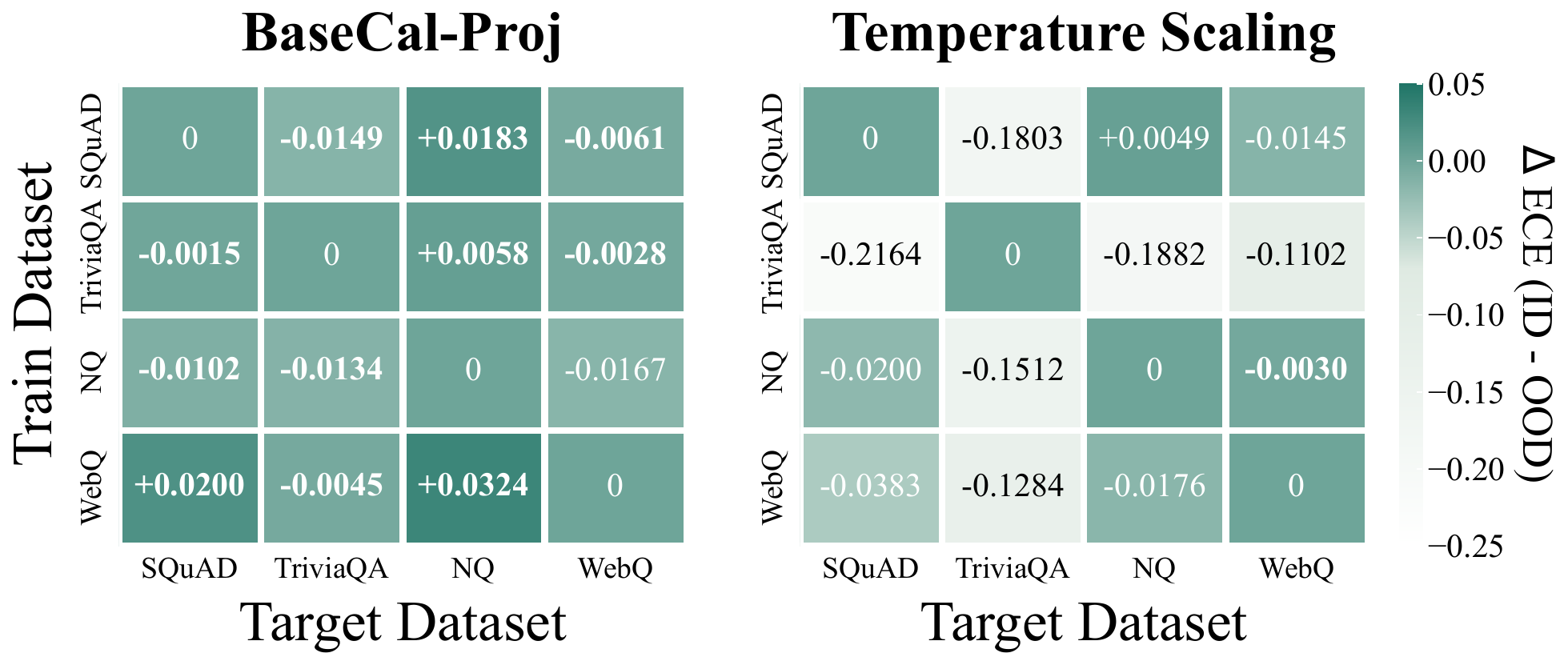}
    \caption{Comparison of $\Delta \text{ECE}$ for BaseCal-Proj and Temperature Scaling for Llama3.1-8b. Darker colors indicate a better out-of-domain performance. The results for other LLMs are shown in Appendix \ref{appendix:generalization}.}
    \label{fig:heatmap_main_text}
\end{figure}

Figure \ref{fig:heatmap_main_text} presents the $\Delta \text{ECE}$ heatmap for BaseCal-Proj and Temperature Scaling. 
Darker colors in the heatmap indicate better OOD calibration performance. 
Overall, \textbf{BaseCal-Proj achieves an average $\Delta \text{ECE}$ of +0.0005, indicating that its OOD calibration performance is practically on par with its in-domain results}.
In contrast, Temperature Scaling suffers from a significant performance degradation with an average $\Delta \text{ECE}$ of -0.0886.
This failure stems from its reliance on post-hoc rescaling to fit the specific correctness label distribution of the training data, rendering it ineffective when accuracy levels shift across datasets (e.g., between TriviaQA and NQ).
In contrast, BaseCal-Proj is label-agnostic and focuses on recovering the intrinsic calibration information embedded within the model's hidden states. 
This prevents overfitting to specific datasets, allowing BaseCal-Proj to maintain consistent performance in unseen questions.

\begin{table}[t]
  \centering
  \begin{adjustbox}{max width=\linewidth}
  \begin{tabular}{ll *{6}{c}}
    \toprule
    \multirow{2.5}{*}{\textbf{Model}} &
    \multirow{2.5}{*}{\textbf{Projection}} &
    \multicolumn{2}{c}{\textbf{TriviaQA}} &
    \multicolumn{2}{c}{\textbf{NQ}} &
    \multicolumn{2}{c}{\textbf{MMLU}} \\
    \cmidrule(lr){3-4} \cmidrule(lr){5-6} \cmidrule(lr){7-8}
    & & ECE ($\downarrow$) & BS ($\downarrow$) & ECE ($\downarrow$) & BS ($\downarrow$) & ECE ($\downarrow$) & BS ($\downarrow$)\\
    \midrule
    Qwen2.5-7B
      & Non-Linear       & 0.0504 & 0.1968 & 0.2813 &	0.2573 & 0.0790 & 0.1633 \\
      -Instruct & Linear       & 0.1393 & 0.1923 & 0.3382	& 0.2880 & 0.0889 & 0.1662 \\
      
    \midrule
    Llama3.1-8B
      & Non-Linear          & 0.1526 & 0.2209 & 0.2052 &	0.2251 & 0.0166 & 0.1487 \\
      -Instruct & Linear      & 0.0387 & 0.1850 & 0.2488	& 0.2358 & 0.0336 & 0.1500 \\
      
    \midrule
    Olmo2-7B
      & Non-Linear          & 0.0783 & 0.2133 & 0.1933 &	0.2228 & 0.0508 & 0.1742 \\
      -Instruct & Linear       & 0.0314 & 0.1966 & 0.2712	& 0.2511 & 0.0525 & 0.1768 \\
     
    \bottomrule
  \end{tabular}
  \end{adjustbox}
  \caption{Performance comparison of BaseCal-Proj with different projection model architectures.}
  \label{table:projection_model}
\end{table}

\subsection{Effect of Projection Architecture (RQ3)}
\label{analyse:projection_model}

Previous experiments adopted a simple linear projection.
Although its effectiveness has been empirically validated, an open question remains as to whether more expressive projection models can further improve calibration performance.
This section additionally evaluates a more complex projection model, a three-layer multilayer perceptron with ReLU non-linearities.
As shown in Table \ref{table:projection_model}, the model with non-linearities achieves a slightly lower average ECE than the linear projection (0.1231 vs.\ 0.1381).
These results suggest that a simple linear projection is already sufficient to effectively recover the calibration, and that increasing the model complexity yields only slight gains.

\begin{table}[t]
    \centering
    \begin{adjustbox}{max width=\linewidth}
    \begin{tabular}{lcccc}
    \toprule
         \multirow{2}{*}{\textbf{Method}} &  \textbf{Olmo2-7b} & \textbf{Olmo2-7b} & \textbf{Olmo2-7b} \\
         & \textbf{-SFT} & \textbf{-DPO} & \textbf{-Instruct} \\
    \midrule
         \textbf{Vanilla}  & 0.1628 & 0.1958 & 0.2121 \\
         \textbf{Verbalization} & 0.3004 & 0.2274 & 0.2054 \\
         \textbf{Semantic Entropy} & 0.2583 &	0.1966	& 0.1910 \\
         \textbf{BaseCal-Proj} &  \textbf{0.0582} & \textbf{0.0269} & \textbf{0.0314} \\
    \bottomrule
    \end{tabular}
    \end{adjustbox}
    \caption{ECE of BaseCal-Proj and baselines across different post-training stages on TriviaQA.}
    \label{tab:result_olmo_stage}
\end{table}

\begin{table}[t] %
\centering
    \begin{adjustbox}{max width=\linewidth}
    \begin{tabular}{lcccccc}
    \toprule
    \multirow{2.5}{*}{\textbf{Scale}} & \multicolumn{3}{c}{\textbf{ECE} ($\downarrow$)} & \multicolumn{3}{c}{\textbf{BS} ($\downarrow$)} \\
    \cmidrule(lr){2-4} \cmidrule(lr){5-7}
    & Vanilla & Proj & ReEval & Vanilla & Proj & ReEval \\
    \midrule
    7B  & 0.3406 & \underline{0.1393} & \textbf{0.1120} & 0.3118 & \underline{0.1923} & \textbf{0.1789} \\
    14B & 0.2687 & \underline{0.0778} & \textbf{0.0663} & 0.2652 & \underline{0.1715} & \textbf{0.1674} \\
    32B & 0.2662 & \underline{0.0854} & \textbf{0.0542} & 0.2514 & \underline{0.1593} & \textbf{0.1623} \\
    72B & 0.2089 & \underline{0.0502} & \textbf{0.0440} & 0.2078 & \underline{0.1433} & \textbf{0.1396} \\
    \bottomrule
    \end{tabular}
\end{adjustbox}
\caption{Calibration performance across different LLM scales of Qwen2.5 instruct series on TriviaQA. \textbf{Proj} and \textbf{ReEval} denote BaseCal-Proj and BaseCal-ReEval.}
\label{tab:scale}
\end{table}

\subsection{Impact of Different PoLLMs (RQ4)}

To answer ``\textbf{RQ4.} How do the proposed methods perform across different PoLLMs?'', we investigate the performance of BaseCal across PoLLMs with varying post-training strategies and scales.

First, we evaluate BaseCal-Proj across different post-training strategies using Olmo2-7B checkpoints, namely Olmo2-7B-SFT/DPO/Instruct, which correspond to the checkpoints after SFT, DPO, and RLVR stages, respectively.         
As shown in Table~\ref{tab:result_olmo_stage}, \textbf{BaseCal-Proj consistently achieves the best performance across all post-training strategies}. 
Notably, verbalization exhibits inconsistent performance across different strategies, even underperforming vanilla for DPO and SFT, due to its heavy reliance on the model's instruction-following capability.
In contrast, BaseCal-Proj effectively recovers well-calibrated confidence agnostic to specific post-training strategies.

Subsequently, the influence of model size is analyzed using Qwen2.5-7b/14b/32b/72b-Instruct. 
Table~\ref{tab:scale} shows that \textbf{BaseCal-Proj consistently yields substantial improvements across all scales}. 
Furthermore, BaseCal-Proj achieves superior performance on larger models, due to the stronger calibration capabilities of larger
base LLMs, a trend also observed in previous work \citep{zhu-etal-2023-calibration}.

\subsection{Application (RQ5)}

\begin{figure}[t]
    \centering
    \includegraphics[width=\linewidth]{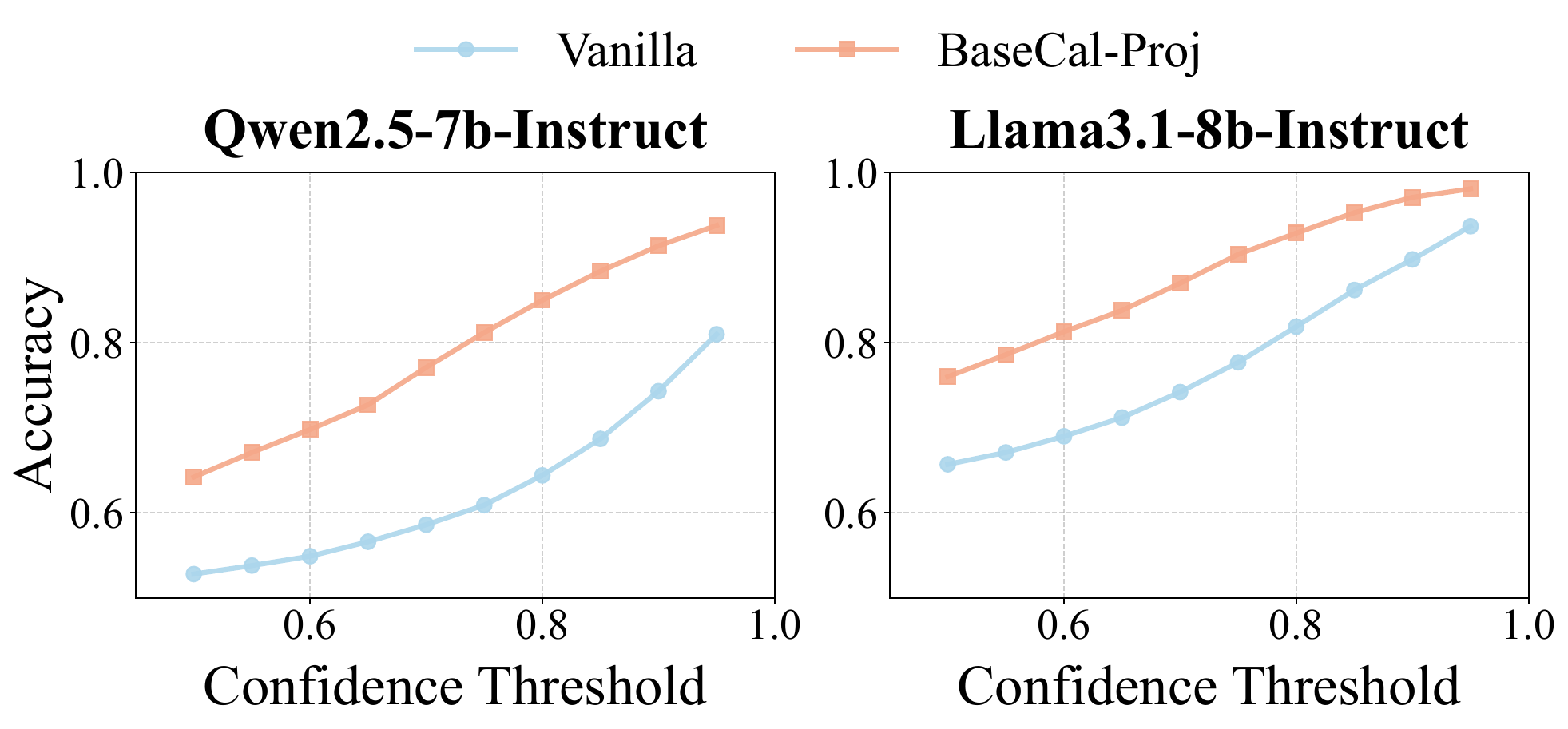}
    \caption{Selective classification accuracy on TriviaQA. Accuracy is computed on examples with confidence scores above thresholds ranging from 0.5 to 0.95.}
    \label{fig:selective_classification}
\end{figure}

Selective classification \citep{selective_classification} enables a model to abstain from making predictions when confidence is low, thereby enhancing reliability by trading off coverage for higher accuracy. 
This mechanism is particularly critical for LLMs in decision-making tasks, where unreliable outputs can lead to severe consequences. 

In this section, we evaluate the effectiveness of our proposed BaseCal-Proj in the context of selective classification. 
\Cref{fig:selective_classification} compares the accuracy of the vanilla baseline and BaseCal-Proj by varying the confidence threshold from 0.5 to 0.95, where prediction below the threshold is rejected.
Experimental results demonstrate that BaseCal-Proj consistently surpasses the vanilla method in accuracy across all confidence thresholds, highlighting its superior capability in identifying reliable predictions.

\section{Conclusion}
To address the overconfidence in PoLLMs, this work first reveals a critical phenomenon where base LLMs retain superior calibration compared to their post-trained counterparts, even on general QA tasks.
Motivated by this, we propose BaseCal, an unsupervised, plug-and-play framework that calibrates PoLLMs with their base counterparts without modifying PoLLM's parameters or compromising generation quality. 
Experiments demonstrate that BaseCal substantially mitigates overconfidence across diverse datasets, LLM size, and post-training strategies.

Beyond performance improvements, our findings provide a key insight into the effect of post-training on confidence calibration. 
The effectiveness of a simple linear projection suggests that calibration information is not entirely lost during post-training. 
Instead, it remains recoverable via a simple linear transformation on the internal states.
By restoring calibration without expensive retraining or human supervision, our method offers a practical solution for deploying reliable LLMs.

\section{Limitations}
This work focuses on providing a practical calibration method based on the observation that base LLMs are relatively well-calibrated, while a theoretical analysis of the underlying mechanisms driving this phenomenon remains an open question.
Additionally, our evaluation primarily targets mainstream factual QA tasks, leaving unexplored long-form generation and complex reasoning tasks, which may introduce task-specific challenges such as uncertainty propagation.
Finally, BaseCal requires access to the base LLM's output layer, which may limit its applicability in scenarios involving closed-source API-based models or environments where internal model parameters are inaccessible.

\section{Ethics Statement}

\paragraph{Data} All data used in this study are publicly available and do not raise any privacy concerns.

\paragraph{AI Writing Assistance} We used ChatGPT solely to refine and polish the textual expressions. 
It was not used to generate new ideas or influence the method design.

\section*{Acknowledgments}

This work was supported by the Strategic Priority Research Program of the Chinese Academy of Sciences (XDB0680201), the Beijing Natural Science Foundation (4252023), and the Key Research and Development Project of Henan Province (No.241111211900).

\bibliography{custom}

\appendix

\section{Appendix}
\label{sec:appendix}

\subsection{Further Information about Datasets}
\label{appendix:datasets}

\begin{table}[h]
\centering
    \begin{adjustbox}{max width=\linewidth}
    \begin{tabular}{lrrrrr}
    \toprule
    Dataset & Train  & Used Train & Dev  & Used Dev & Test \\
    \midrule
    TriviaQA    & 70,098 & 10000 &  17,524 &  2000 & 11,313 \\
    SQuAD       & 70,079 & 10000 & 17,520 & 2000 &10,570 \\
    NQ          & 79,168 & 10000 & 8,757 & 2000 & 3,610  \\
    WebQ & 3,022 & 3022 & 756  & 756 & 2,032  \\
    MMLU        & 98,843 & 10000 & 10,000 & 2000 & 14,042 \\
    \bottomrule
    \end{tabular}
    \end{adjustbox}
\caption{Statistics of the datasets. ``Used Train'' and ``Used Dev'' show the number of data used to train or validate our BaseCal-Proj method.}
\label{tab:dataset_stats}
\end{table}

\subsubsection{Statistics}
\Cref{tab:dataset_stats} shows the detailed statistics of the datasets.
To construct the training data for BaseCal-Proj, we randomly sample 10,000 questions from the original training sets, as this scale suffices for our lightweight linear projection model. 
To mitigate the risk of overfitting, we further sample 2,000 questions from the dev set for validation and employ an early stopping strategy based on the validation loss. 
For the WebQ dataset, which contains fewer instances, we utilize all 3,022 available questions for training and 756 for validation. 
Crucially, during both the training and validation phases, our method relies solely on the input questions without accessing any human-labeled ground truth.

\subsubsection{Generation Prompts}
In this section, we present the prompts used for response generation across the different datasets.
For the QA datasets, including TriviaQA, SQuAD, WebQ, and NQ, we employ the following prompt template:
\begin{quote}
    \ttfamily
    Answer the question briefly. \{5 examples\}\\
    Question: \{query\}\\
    Answer:
\end{quote}

For the MMLU multiple-choice dataset, we utilize the following prompt to elicit answers and extract the option labels (A, B, C, or D):
\begin{quote}
    \ttfamily
    Answer the following multiple-choice question. Reply with only A, B, C, or D. \{5 examples\}\\
    Question: \{question\}\\
    A. \{choices[0]\} \quad B. \{choices[1]\}\\
    C. \{choices[2]\} \quad D. \{choices[3]\} \\
    Answer: 
\end{quote}

\subsection{Details about LLM-as-a-Judge}
\label{appendix:llm-as-a-judge}
To assess the correctness of model outputs in QA tasks, we employ an LLM-as-a-judge to determine whether the generated responses are semantically equivalent to the ground truth, following \citet{tian-etal-2023-just, simpleqa}. 
For reproducibility, we utilize the powerful open-source Qwen2.5-14B-Instruct model as our judge. 
Inspired by the evaluation protocol from \citet{simpleqa}, we use the specific prompt detailed in \Cref{section:prompt_llm-as-a-judge} to verify response correctness. 
To validate the quality of the LLM judgment, we conducted a manual audit on 100 randomly sampled instances. Comparative verification against human ground truth reveals a disagreement rate of only 1\% with human annotations, confirming the reliability of our automated evaluation process.

\subsection{Further Information about Baselines}
\label{appendix:baselines}

This section provides further details for the calibration baselines.

\textbf{(1)} \textbf{Vanilla}. Several studies have employed the aggregated token probabilities as the vanilla confidence \citep{orgad2024llmsknowshowintrinsic, factualconfidence, malinin2021uncertainty}. 
Following implementation in  \citet{orgad2024llmsknowshowintrinsic}, we average the probabilities of all tokens in a response as the confidence.

\textbf{(2)} \textbf{Verbalization}. This method leverages the instruction-following capability of models to express confidence in natural language. Following the implementation of \citet{tian-etal-2023-just}, we employ the following prompt:
\begin{quote}
    \ttfamily
    Provide the probability that your guess is correct. Give ONLY the probability, no other words or explanation. For example:\\
    Probability: <the probability between 0.0 and 1.0 that your guess is correct, without any extra commentary whatsoever; just the probability!>\\
    The question is: \{question\} \\
    The best guess is: \{response\}\\
    Probability: 
\end{quote}
The resulting output is then parsed using regular expressions to extract a floating-point value between 0.0 and 1.0, which serves as the confidence score.
Notably, this method requires an additional forward pass on the PoLLMs after the initial response is generated, thereby increasing the overall computational overhead.

\textbf{(3)} \textbf{P(True)}. This method follows the prompting strategy introduced by \citet{kadavath2022languagemodelsmostlyknow}, where the LLM is directly queried to assess the correctness of its own output. 
P(True) also requires an additional forward pass on the PoLLMs after the initial response is generated.
Specifically, we construct the following prompt:
\begin{quote}
    \ttfamily
    Question: \{question\} \\
    Possible answer: \{response\} \\
    Is the possible answer: \\A. True \quad B. False \\
    The possible answer is:
\end{quote}
The model's confidence is then quantified as the probability it assigns to the token ``A''.

\textbf{(4)} \textbf{Semantic Entropy}. As proposed by \citet{semantic_entropy, farquhar2024detecting}, semantic entropy estimates confidence by sampling multiple responses for the same question and computing their semantic consistency. Following the implementation details recommended by \citet{semantic_entropy}, we set the sampling parameters as follows: temperature 0.5, number of samples 10, top\_p = 1.0, and top\_k = -1. Since this metric was originally designed for hallucination detection, its raw scores do not fall within the $[0,1]$ interval. Therefore, following \citet{li-etal-2025-towards}, we normalize the semantic entropy values to the range $[0,1]$ to facilitate the calculation of ECE.

\begin{figure*}[t]
    \centering
    \includegraphics[width=0.9\linewidth]{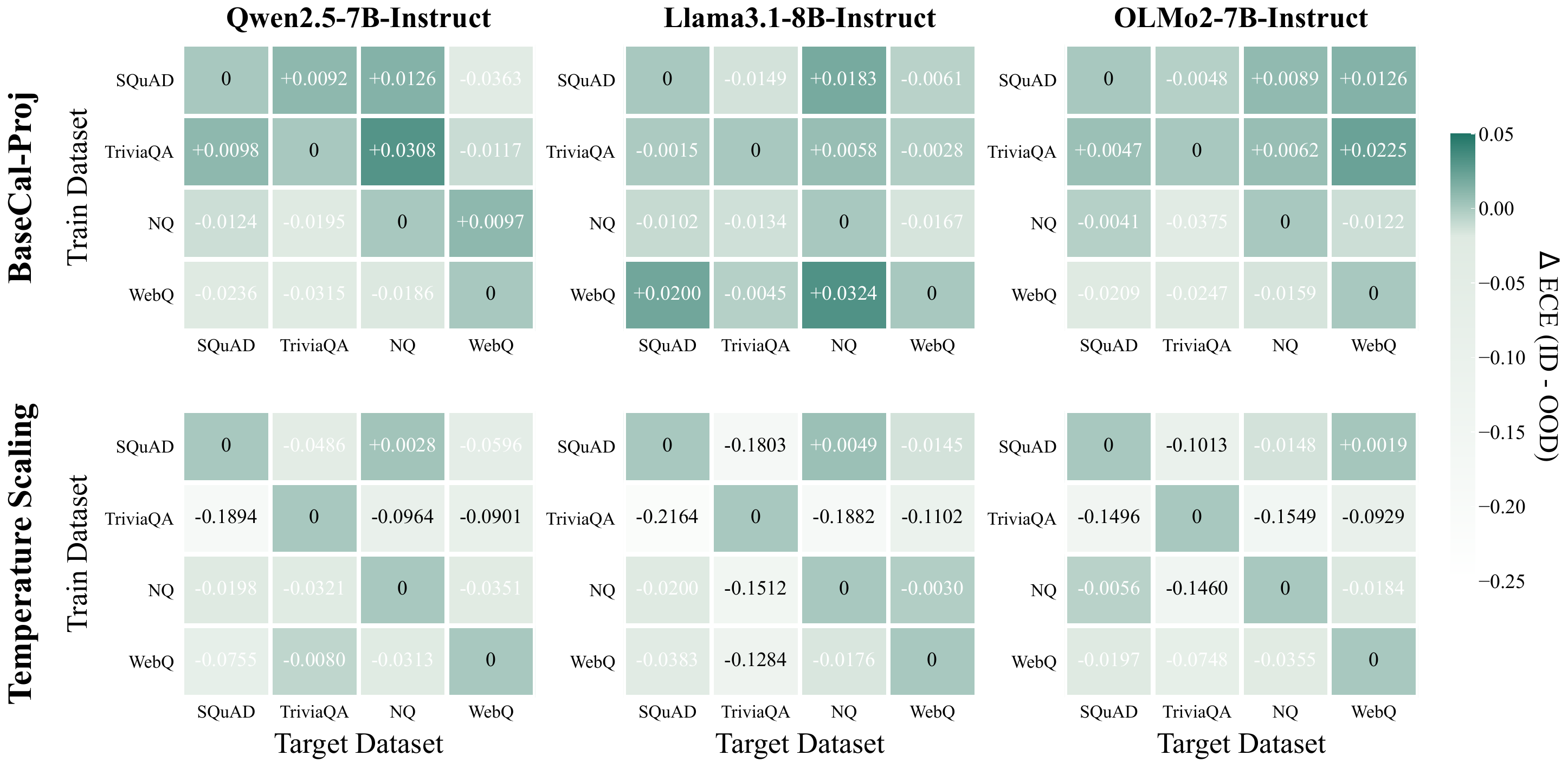}
    \caption{Comparison of $\Delta \text{ECE}$ for BaseCal-Proj and Temperature Scaling. Darker colors indicate a better out-of-domain performance.}
    \label{fig:heatmap_appendix}
\end{figure*}

\begin{table}[t]
  \centering
  \begin{adjustbox}{max width=\linewidth}
  \begin{tabular}{ll *{6}{c}}
    \toprule
    \multirow{2}{*}{\textbf{Model}} &
    \multirow{2}{*}{\textbf{Loss}} &
    \multicolumn{2}{c}{\textbf{TriviaQA}} &
    \multicolumn{2}{c}{\textbf{NQ}} &
    \multicolumn{2}{c}{\textbf{MMLU}} \\
    \cmidrule(lr){3-4} \cmidrule(lr){5-6} \cmidrule(lr){7-8}
    & & ECE ($\downarrow$) & BS ($\downarrow$) & ECE ($\downarrow$) & BS ($\downarrow$) & ECE ($\downarrow$) & BS ($\downarrow$)\\
    \midrule
    \multirow{3}{*}{\makecell[l]{Qwen2.5-7B\\-Instruct}}
       & MSE       & 0.1393 & 0.1923 & 0.3382	& 0.2880 & 0.0889 & 0.1662 \\
       & MAE & 0.1260 & 0.1898 & 0.3276 & 0.2806 &  0.0882 & 0.1665 \\
       & Cosine & 0.5011 & 0.5009 & 0.2543 & 0.2548 & 0.0396 & 0.1592 \\
    \midrule
    \multirow{3}{*}{\makecell[l]{Llama3.1-8B\\-Instruct}}
      & MSE       & 0.0387 & 0.1850 & 0.2488	& 0.2358 & 0.0336 & 0.1500 \\
        & MAE & 0.0447 & 0.1854 & 0.2421 & 0.2339 & 0.0332 & 0.1500 \\
      & Cosine & 0.6125 & 0.6057 & 0.2483 & 0.2597 & 0.0221 & 0.1489 \\
    \midrule
    \multirow{3}{*}{\makecell[l]{Olmo2-7B\\-Instruct}}
    & MSE       & 0.0314 & 0.1966 & 0.2712	& 0.2511 & 0.0525 & 0.1768 \\
    & MAE & 0.0297 & 0.1969 & 0.2676 & 0.2493 & 0.0587 & 0.1768 \\
    & Cosine & 0.5651 & 0.5636 & 0.2929 & 0.2935 & 0.0384 & 0.1718 \\
    \bottomrule
  \end{tabular}
  \end{adjustbox}
  \caption{Performance comparison of BaseCal-Proj trained with different loss functions.}
  \label{table:loss}
\end{table}

\textbf{(5)} \textbf{DACA}. DACA \citep{luo2025your} is an unsupervised approach that optimizes a temperature parameter to scale the probability of PoLLM, thereby approximating the probability distribution of the base LLM. 
This probability-level scaling necessitates that both the base LLM and PoLLM produce the \textbf{exact same top-1 token} for a given input, a constraint theoretically proved and empirically analyzed in Section 3 (pp. 4-5) of \citet{luo2025your}. 
This limitation restricts DACA's applicability primarily to multiple-choice tasks, as identical token generation is very rare for base LLM and PoLLM in free-form QA. 
Accordingly, we compare our method with DACA on the MMLU benchmark in the main experiments. 
We utilize the official implementation \citep{luo2025your} and optimize DACA using the same training set employed for our proposed method.

\textbf{(6)} \textbf{Temperature Scaling}.
We directly adopt the implementation of temperature scaling from \citet{xia-etal-2025-influences}. 
This is a supervised method that optimizes an optimal temperature parameter to rescale probabilities, thereby minimizing the negative log-likelihood. 
The training process requires questions and human-labeled answers to determine response correctness.
In our implementation, we utilize the entire training dataset to optimize the temperature.

\subsection{Further Information about BaseCal-Proj}

\subsubsection{Effect of Loss Function}
\label{appendix:loss}

The default BaseCal-Proj optimization utilizes the Mean Squared Error (MSE) to minimize the Euclidean distance between the projected PoLLM states and the target states:
$$
\mathcal{L}_{\text{MSE}}(\theta)
=
\frac{1}{T}\sum_{t=1}^{T}
\left\|
\phi_{\theta}\!\left(\boldsymbol{h}^{\text{p}}_{t-1}\right)
-
\boldsymbol{h}^{\text{b}}_{t-1}
\right\|_2^2.
$$

To assess the effect of different loss functions, we explore two alternative objectives. First, we consider the Mean Absolute Error (MAE), which imposes an $L_1$ penalty:
$$
\mathcal{L}_{\text{MAE}}(\theta)
=
\frac{1}{T}\sum_{t=1}^{T}
\left\|
\phi_{\theta}\!\left(\boldsymbol{h}^{\text{p}}_{t-1}\right)
-
\boldsymbol{h}^{\text{b}}_{t-1}
\right\|_1.
$$

Second, we examine the Cosine Loss, which prioritizes angular alignment over magnitude:
$$
\mathcal{L}_{\text{Cos}}(\theta)
=
\frac{1}{T}\sum_{t=1}^{T}
\left(
1 -
\frac{
\phi_{\theta}\!\left(\boldsymbol{h}^{\text{p}}_{t-1}\right)^\top \boldsymbol{h}^{\text{b}}_{t-1}
}{
\left\| \phi_{\theta}\!\left(\boldsymbol{h}^{\text{p}}_{t-1}\right) \right\|_2 \left\| \boldsymbol{h}^{\text{b}}_{t-1} \right\|_2
}
\right).
$$

To ensure a fair comparison, we maintain a consistent architecture using a single linear layer for all projection models.
\Cref{table:loss} shows the calibration performance of BaseCal-Proj trained with MSE, MAE, and Cosine Loss.
The experimental results reveal that $\mathcal{L}_{\text{Cos}}$ exhibits significant instability across different benchmarks. While it achieves competitive calibration on MMLU, its performance degrades sharply on TriviaQA.
This discrepancy suggests that solely optimizing the angular alignment of hidden states is insufficient for confidence calibration.
In contrast, both MSE and MAE demonstrate consistently superior performance across all tested models and datasets.

\subsubsection{More Results about Generalization}
\label{appendix:generalization}

Figure \ref{fig:heatmap_appendix} illustrates the generalization gap, $\Delta \text{ECE}$, for the Qwen, Llama, and Olmo models. 
It can be observed that across all model-dataset combinations, the $\Delta \text{ECE}$ for our method is close to 0, demonstrating that BaseCal-Proj achieves OOD performance comparable to its ID performance. 
In contrast, temperature scaling generally exhibits significantly lower $\Delta \text{ECE}$ values, reaching as low as -0.2164. 
This indicates a significant degradation in the performance of temperature scaling when generalizing across different datasets.

\subsection{Prompt for LLM-as-a-Judge}
\label{section:prompt_llm-as-a-judge}

\onecolumn
\begin{tcolorbox}[
  colback=gray!5,
  colframe=gray!75!black,
  title=Prompt for LLM-as-a-Judge,
  fonttitle=\bfseries,
  breakable,
  width=\textwidth,
  left=3pt,
  right=3pt,
  top=3pt,
  bottom=3pt,
  boxsep=2pt,
]
{\footnotesize
\begin{lstlisting}[breaklines=true, columns=flexible]
Your job is to look at a question, some gold targets, and a predicted answer, and then assign a grade of either ["CORRECT", "INCORRECT", "NOT_ATTEMPTED"].
First, I will give examples of each grade, and then you will grade a new example.

The following are examples of CORRECT predicted answers.

Question: What are the names of Barack Obama's children?
Gold target: ["Malia Obama and Sasha Obama", "malia and sasha"]
Predicted answer 1: sasha and malia obama
Predicted answer 2: most people would say Malia and Sasha, but I'm not sure and would have to double check
Predicted answer 3: Barack Obama has two daughters. Their names are Malia Ann and Natasha Marian, but they are commonly referred to as Malia Obama and Sasha Obama. Malia was born on July 4, 1998, and Sasha was born on June 10, 2001.

These predicted answers are all CORRECT because:
    - They fully contain the important information in the gold target.
    - They do not contain any information that contradicts the gold target.
    - Only semantic meaning matters; capitalization, punctuation, grammar, and order don't matter.
    - Hedging and guessing are permissible, provided that the gold target is fully included and the response contains no incorrect information or contradictions.


The following are examples of INCORRECT predicted answers.

Question: What are the names of Barack Obama's children?
Gold target: ["Malia and Sasha"]
Predicted answer 1: Malia.
Predicted answer 2: Malia, Sasha, and Susan.
Predicted answer 3: Barack Obama does not have any children.
Predicted answer 4: I think it's either Malia and Sasha. Or it could be Malia and Jackie. Or it could be Joey and Malia.
Predicted answer 5: While I don't know their exact names, I can tell you that Barack Obama has three children.
Predicted answer 6: It's possible you may mean Betsy and Olivia. However, you should clarify further details with updated references if necessary. Is that the correct answer?
Predicted answer 7: It may be the case that Obama's child is named James. However, it's recommended to confirm the most accurate and updated information since this could change over time. This model may not always reflect the most current information.

These predicted answers are all INCORRECT because:
    - A factual statement in the answer contradicts the gold target. Incorrect statements that have some hedging (e.g., "it is possible that", "although i'm not sure, i think") are also considered incorrect.


The following are examples of NOT_ATTEMPTED predicted answers.

Question: What are the names of Barack Obama's children?
Gold target: ["Malia and Sasha"]
Predicted answer 1: I don't know.
Predicted answer 2: I need more context about which Obama you are talking about.
Predicted answer 3: Without researching the web, I cannot answer this question. However, I can tell you that Barack Obama has two children.
Predicted answer 4: Barack Obama has two children. I know that one of them is Malia, but I'm not sure about the other one.

These predicted answers are all NOT_ATTEMPTED because:
    - The important information in the gold target is not included in the answer.
    - No statements in the answer contradict the gold target.


Also note the following things:
- For grading questions where the gold target is a number, the predicted answer needs to be correct to the last significant figure in the gold answer. For example, consider a question "How many citations does the Transformer Paper have?" with gold target "120k". 
    - Predicted answers "120k", "124k", and "115k" are all CORRECT. 
    - Predicted answers "100k" and "113k" are INCORRECT. 
    - Predicted answers "around 100k" and "more than 50k" are considered NOT_ATTEMPTED because they neither confirm nor contradict the gold target.
- The gold target may contain more information than the question. In such cases, the predicted answer only needs to contain the information that is in the question.
    - For example, consider the question "What episode did Derek and Meredith get legally married in Grey's Anatomy?" with gold target "Season 7, Episode 20: White Wedding". Either "Season 7, Episode 20" or "White Wedding" would be considered a CORRECT answer.
- Do not punish predicted answers if they omit information that would be clearly inferred from the question.
    - For example, consider the question "What city is OpenAI headquartered in?" and the gold target "San Francisco, California". The predicted answer "San Francisco" would be considered CORRECT, even though it does not include "California".
    - Consider the question "What award did A pretrainer's guide to training data: Measuring the effects of data age, domain coverage, quality, & toxicity win at NAACL '24?", the gold target is "Outstanding Paper Award". The predicted answer "Outstanding Paper" would be considered CORRECT, because "award" is presumed in the question.
    - For the question "What is the height of Jason Wei in meters?", the gold target is "1.73 m". The predicted answer "1.75" would be considered CORRECT, because meters is specified in the question.
    - For the question "What is the name of Barack Obama's wife?", the gold target is "Michelle Obama". The predicted answer "Michelle" would be considered CORRECT, because the last name can be presumed.
- Do not punish for typos in people's name if it's clearly the same name. 
    - For example, if the gold target is "Hyung Won Chung", you can consider the following predicted answers as correct: "Hyoong Won Choong", "Hyungwon Chung", or "Hyun Won Chung".


Here is a new example. Simply reply with either CORRECT, INCORRECT, NOT_ATTEMPTED. Don't apologize or correct yourself if there was a mistake; we are just trying to grade the answer.

Question: {question}
Gold target: {target}
Predicted answer: {predicted_answer}

Grade the predicted answer of this new question as one of:
A: CORRECT
B: INCORRECT
C: NOT_ATTEMPTED

Just return the letters "A", "B", or "C", with no text around it.
\end{lstlisting}
}
\end{tcolorbox}
\twocolumn

\end{document}